\documentclass{article}

\PassOptionsToPackage{numbers, sort}{natbib}
\usepackage[main, final]{neurips_2026}   %

\usepackage[utf8]{inputenc} %
\usepackage[T1]{fontenc}    %
\usepackage{hyperref}       %
\hypersetup{hidelinks, pdftitle={VolFill: Single-View Amodal 3D Scene Reconstruction with Volumetric Flow Matching}, pdfauthor={Tuan Duc Ngo, Chuang Gan, Evangelos Kalogerakis}}
\usepackage{url}            %
\usepackage{booktabs}       %
\usepackage{amsfonts}       %
\usepackage{nicefrac}       %
\usepackage{microtype}      %
\usepackage{graphicx}
\usepackage{arydshln}
\usepackage{multirow}
\usepackage{caption}
\usepackage{comment}
\usepackage[table]{xcolor}
\usepackage{colortbl} %
\usepackage{xspace}
\usepackage{subfigure}
\usepackage[symbol]{footmisc}
\usepackage{enumitem}
\usepackage{amsmath,amsfonts,bm}
\usepackage{comment}
\usepackage[normalem]{ulem}

\usepackage{threeparttable} %

\usepackage{amssymb}%
\usepackage{pifont}%
\newcommand{\xmark}{\ding{55}}%

\definecolor{custom_red}{RGB}{231,111,81}
\definecolor{custom_green}{RGB}{42,157,143}
\definecolor{custom_dark}{RGB}{38,70,83}
\definecolor{custom_yellow}{RGB}{233,196,106}
\definecolor{custom_orange}{RGB}{244,162,97}

\title{\Approach: Single-View Amodal 3D Scene Reconstruction with Volumetric Flow Matching}

\definecolor{accentgrad}{HTML}{514FEC}

\author{%
  Tuan Duc Ngo$^{1}$ \quad Chuang Gan$^{1}$ \quad Evangelos Kalogerakis$^{1,2}$\\[2pt]
  $^{1}$UMass Amherst \qquad $^{2}$Technical University of Crete \\[4pt]
  \href{https://ngoductuanlhp.github.io/VolFill/}{\textcolor{accentgrad}{
  \texttt{ngoductuanlhp.github.io/VolFill}}
  }
}

\begin{document}

\def\mA{\mathcal{A}}
\def\mB{\mathcal{B}}
\def\mC{\mathcal{C}}
\def\mD{\mathcal{D}}
\def\mE{\mathcal{E}}
\def\mF{\mathcal{F}}
\def\mG{\mathcal{G}}
\def\mH{\mathcal{H}}
\def\mI{\mathcal{I}}
\def\mJ{\mathcal{J}}
\def\mK{\mathcal{K}}
\def\mL{\mathcal{L}}
\def\mM{\mathcal{M}}
\def\mN{\mathcal{N}}
\def\mO{\mathcal{O}}
\def\mP{\mathcal{P}}
\def\mQ{\mathcal{Q}}
\def\mR{\mathcal{R}}
\def\mS{\mathcal{S}}
\def\mT{\mathcal{T}}
\def\mU{\mathcal{U}}
\def\mV{\mathcal{V}}
\def\mW{\mathcal{W}}
\def\mX{\mathcal{X}}
\def\mY{\mathcal{Y}}
\def\mZ{\mathcal{Z}} 

\def\bbN{\mathbb{N}} 
\def\bbR{\mathbb{R}} 
\def\bbP{\mathbb{P}} 
\def\bbQ{\mathbb{Q}} 
\def\bbE{\mathbb{E}}

\def\1n{\mathbf{1}_n}
\def\0{\mathbf{0}}
\def\1{\mathbf{1}}

\def\A{{\bf A}}
\def\B{{\bf B}}
\def\C{{\bf C}}
\def\D{{\bf D}}
\def\E{{\bf E}}
\def\F{{\bf F}}
\def\G{{\bf G}}
\def\H{{\bf H}}
\def\I{{\bf I}}
\def\J{{\bf J}}
\def\K{{\bf K}}
\def\L{{\bf L}}
\def\M{{\bf M}}
\def\N{{\bf N}}
\def\O{{\bf O}}
\def\P{{\bf P}}
\def\Q{{\bf Q}}
\def\R{{\bf R}}
\def\S{{\bf S}}
\def\T{{\bf T}}
\def\U{{\bf U}}
\def\V{{\bf V}}
\def\W{{\bf W}}
\def\X{{\bf X}}
\def\Y{{\bf Y}}
\def\Z{{\bf Z}}

\def\a{{\bf a}}
\def\b{{\bf b}}
\def\c{{\bf c}}
\def\d{{\bf d}}
\def\e{{\bf e}}
\def\f{{\bf f}}
\def\g{{\bf g}}
\def\h{{\bf h}}
\def\i{{\bf i}}
\def\j{{\bf j}}
\def\k{{\bf k}}
\def\l{{\bf l}}
\def\m{{\bf m}}
\def\n{{\bf n}}
\def\o{{\bf o}}
\def\p{{\bf p}}
\def\q{{\bf q}}
\def\r{{\bf r}}
\def\s{{\bf s}}
\def\t{{\bf t}}
\def\u{{\bf u}}
\def\v{{\bf v}}
\def\w{{\bf w}}
\def\x{{\bf x}}
\def\y{{\bf y}}
\def\z{{\bf z}}

\def\balpha{\mbox{\boldmath{$\alpha$}}}
\def\bbeta{\mbox{\boldmath{$\beta$}}}
\def\bdelta{\mbox{\boldmath{$\delta$}}}
\def\bgamma{\mbox{\boldmath{$\gamma$}}}
\def\blambda{\mbox{\boldmath{$\lambda$}}}
\def\bsigma{\mbox{\boldmath{$\sigma$}}}
\def\btheta{\mbox{\boldmath{$\theta$}}}
\def\bomega{\mbox{\boldmath{$\omega$}}}
\def\bxi{\mbox{\boldmath{$\xi$}}}
\def\bnu{\mbox{\boldmath{$\nu$}}}                                  
\def\bphi{\mbox{\boldmath{$\phi$}}}
\def\bmu{\mbox{\boldmath{$\mu$}}}

\def\bDelta{\mbox{\boldmath{$\Delta$}}}
\def\bOmega{\mbox{\boldmath{$\Omega$}}}
\def\bPhi{\mbox{\boldmath{$\Phi$}}}
\def\bLambda{\mbox{\boldmath{$\Lambda$}}}
\def\bSigma{\mbox{\boldmath{$\Sigma$}}}
\def\bGamma{\mbox{\boldmath{$\Gamma$}}}
                                  
\newcommand{\myprob}[1]{\mathop{\mathbb{P}}_{#1}}

\newcommand{\myexp}[1]{\mathop{\mathbb{E}}_{#1}}

\newcommand{\mydelta}[1]{1_{#1}}

\newcommand{\myminimum}[1]{\mathop{\textrm{minimum}}_{#1}}
\newcommand{\mymaximum}[1]{\mathop{\textrm{maximum}}_{#1}}    
\newcommand{\mymin}[1]{\mathop{\textrm{minimize}}_{#1}}
\newcommand{\mymax}[1]{\mathop{\textrm{maximize}}_{#1}}
\newcommand{\mymins}[1]{\mathop{\textrm{min.}}_{#1}}
\newcommand{\mymaxs}[1]{\mathop{\textrm{max.}}_{#1}}  
\newcommand{\myargmin}[1]{\mathop{\textrm{argmin}}_{#1}} 
\newcommand{\myargmax}[1]{\mathop{\textrm{argmax}}_{#1}} 
\newcommand{\myst}{\textrm{s.t. }}

\newcommand{\denselist}{\itemsep -1pt}
\newcommand{\sparselist}{\itemsep 1pt}

\definecolor{pink}{rgb}{0.9,0.5,0.5}
\definecolor{purple}{rgb}{0.5, 0.4, 0.8}   
\definecolor{gray}{rgb}{0.3, 0.3, 0.3}
\definecolor{mygreen}{rgb}{0.2, 0.6, 0.2}  

\newcommand{\cyan}[1]{\textcolor{cyan}{#1}}
\newcommand{\red}[1]{\textcolor{red}{#1}}  
\newcommand{\blue}[1]{\textcolor{blue}{#1}}
\newcommand{\magenta}[1]{\textcolor{magenta}{#1}}
\newcommand{\pink}[1]{\textcolor{pink}{#1}}
\newcommand{\green}[1]{\textcolor{green}{#1}} 
\newcommand{\gray}[1]{\textcolor{gray}{#1}}    
\newcommand{\mygreen}[1]{\textcolor{mygreen}{#1}}    
\newcommand{\purple}[1]{\textcolor{purple}{#1}}       

\definecolor{greena}{rgb}{0.4, 0.5, 0.1}
\newcommand{\greena}[1]{\textcolor{greena}{#1}}

\definecolor{bluea}{rgb}{0, 0.4, 0.6}
\newcommand{\bluea}[1]{\textcolor{bluea}{#1}}
\definecolor{reda}{rgb}{0.6, 0.2, 0.1}
\newcommand{\reda}[1]{\textcolor{reda}{#1}}

\def\changemargin#1#2{\list{}{\rightmargin#2\leftmargin#1}\item[]}
\let\endchangemargin=\endlist
                                               
\newcommand{\cm}[1]{}

\newcommand{\mhoai}[1]{{\color{magenta}\textbf{[MH: #1]}}}

\newcommand{\mtodo}[1]{{\color{red}$\blacksquare$\textbf{[TODO: #1]}}}
\newcommand{\myheading}[1]{\vspace{1ex}\noindent \textbf{#1}}
\newcommand{\htimesw}[2]{\mbox{$#1$$\times$$#2$}}

\newif\ifshowsolution
\showsolutiontrue

\ifshowsolution  
\newcommand{\Solution}[2]{\paragraph{\bf $\bigstar $ SOLUTION:} {\sf #2} }
\newcommand{\Mistake}[2]{\paragraph{\bf $\blacksquare$ COMMON MISTAKE #1:} {\sf #2} \bigskip}
\else
\newcommand{\Solution}[2]{\vspace{#1}}
\fi

\newcommand{\truefalse}{
\begin{enumerate}
	\item True
	\item False
\end{enumerate}
}

\newcommand{\yesno}{
\begin{enumerate}
	\item Yes
	\item No
\end{enumerate}
}

\newcommand{\Sref}[1]{Sec.~\ref{#1}}
\newcommand{\Eref}[1]{Eq.~(\ref{#1})}
\newcommand{\Fref}[1]{Fig.~\ref{#1}}
\newcommand{\Tref}[1]{Table~\ref{#1}}

\newcommand\blfootnote[1]{%
  \begingroup
  \renewcommand\thefootnote{}\footnote{#1}%
  \addtocounter{footnote}{-1}%
  \endgroup
}

\makeatletter
\DeclareRobustCommand\onedot{\futurelet\@let@token\@onedot}
\def\@onedot{\ifx\@let@token.\else.\null\fi\xspace}

\newcommand{\eg}{\emph{e.g}\onedot}
\newcommand{\Eg}{\emph{E.g}\onedot}
\newcommand{\ie}{\emph{i.e}\onedot}
\newcommand{\Ie}{\emph{I.e}\onedot}
\newcommand{\cf}{\emph{cf}\onedot}
\newcommand{\Cf}{\emph{Cf}\onedot}
\newcommand{\etc}{\emph{etc}\onedot}
\newcommand{\wrt}{w.r.t\onedot}
\newcommand{\dof}{d.o.f\onedot}
\newcommand{\iid}{i.i.d\onedot}
\newcommand{\wolog}{w.l.o.g\onedot}
\newcommand{\etal}{\emph{et al}\onedot}

\def\paperID{xxxxx} %
\def\confName{NIPS}
\def\confYear{2026}

\def\Approach{VolFill\xspace}

\definecolor{graybg}{gray}{0.75} %

\definecolor{custom_red}{RGB}{231,111,81}
\definecolor{custom_green}{RGB}{42,157,143}
\definecolor{custom_dark}{RGB}{38,70,83}
\definecolor{custom_yellow}{RGB}{233,196,106}
\definecolor{custom_orange}{RGB}{244,162,97}

\newcommand{\figleft}{{\em (Left)}}
\newcommand{\figcenter}{{\em (Center)}}
\newcommand{\figright}{{\em (Right)}}
\newcommand{\figtop}{{\em (Top)}}
\newcommand{\figbottom}{{\em (Bottom)}}
\newcommand{\captiona}{{\em (a)}}
\newcommand{\captionb}{{\em (b)}}
\newcommand{\captionc}{{\em (c)}}
\newcommand{\captiond}{{\em (d)}}

\newcommand{\newterm}[1]{{\bf #1}}

\def\figref#1{figure~\ref{#1}}
\def\Figref#1{Figure~\ref{#1}}
\def\twofigref#1#2{figures \ref{#1} and \ref{#2}}
\def\quadfigref#1#2#3#4{figures \ref{#1}, \ref{#2}, \ref{#3} and \ref{#4}}
\def\secref#1{section~\ref{#1}}
\def\Secref#1{Section~\ref{#1}}
\def\twosecrefs#1#2{sections \ref{#1} and \ref{#2}}
\def\secrefs#1#2#3{sections \ref{#1}, \ref{#2} and \ref{#3}}
\def\eqref#1{equation~\ref{#1}}
\def\Eqref#1{Equation~\ref{#1}}
\def\plaineqref#1{\ref{#1}}
\def\chapref#1{chapter~\ref{#1}}
\def\Chapref#1{Chapter~\ref{#1}}
\def\rangechapref#1#2{chapters\ref{#1}--\ref{#2}}
\def\algref#1{algorithm~\ref{#1}}
\def\Algref#1{Algorithm~\ref{#1}}
\def\twoalgref#1#2{algorithms \ref{#1} and \ref{#2}}
\def\Twoalgref#1#2{Algorithms \ref{#1} and \ref{#2}}
\def\partref#1{part~\ref{#1}}
\def\Partref#1{Part~\ref{#1}}
\def\twopartref#1#2{parts \ref{#1} and \ref{#2}}

\def\ceil#1{\lceil #1 \rceil}
\def\floor#1{\lfloor #1 \rfloor}
\def\1{\bm{1}}
\newcommand{\train}{\mathcal{D}}
\newcommand{\valid}{\mathcal{D_{\mathrm{valid}}}}
\newcommand{\test}{\mathcal{D_{\mathrm{test}}}}

\def\eps{{\epsilon}}

\def\reta{{\textnormal{$\eta$}}}
\def\ra{{\textnormal{a}}}
\def\rb{{\textnormal{b}}}
\def\rc{{\textnormal{c}}}
\def\rd{{\textnormal{d}}}
\def\re{{\textnormal{e}}}
\def\rf{{\textnormal{f}}}
\def\rg{{\textnormal{g}}}
\def\rh{{\textnormal{h}}}
\def\ri{{\textnormal{i}}}
\def\rj{{\textnormal{j}}}
\def\rk{{\textnormal{k}}}
\def\rl{{\textnormal{l}}}
\def\rn{{\textnormal{n}}}
\def\ro{{\textnormal{o}}}
\def\rp{{\textnormal{p}}}
\def\rq{{\textnormal{q}}}
\def\rr{{\textnormal{r}}}
\def\rs{{\textnormal{s}}}
\def\rt{{\textnormal{t}}}
\def\ru{{\textnormal{u}}}
\def\rv{{\textnormal{v}}}
\def\rw{{\textnormal{w}}}
\def\rx{{\textnormal{x}}}
\def\ry{{\textnormal{y}}}
\def\rz{{\textnormal{z}}}

\def\rvepsilon{{\mathbf{\epsilon}}}
\def\rvtheta{{\mathbf{\theta}}}
\def\rva{{\mathbf{a}}}
\def\rvb{{\mathbf{b}}}
\def\rvc{{\mathbf{c}}}
\def\rvd{{\mathbf{d}}}
\def\rve{{\mathbf{e}}}
\def\rvf{{\mathbf{f}}}
\def\rvg{{\mathbf{g}}}
\def\rvh{{\mathbf{h}}}
\def\rvu{{\mathbf{i}}}
\def\rvj{{\mathbf{j}}}
\def\rvk{{\mathbf{k}}}
\def\rvl{{\mathbf{l}}}
\def\rvm{{\mathbf{m}}}
\def\rvn{{\mathbf{n}}}
\def\rvo{{\mathbf{o}}}
\def\rvp{{\mathbf{p}}}
\def\rvq{{\mathbf{q}}}
\def\rvr{{\mathbf{r}}}
\def\rvs{{\mathbf{s}}}
\def\rvt{{\mathbf{t}}}
\def\rvu{{\mathbf{u}}}
\def\rvv{{\mathbf{v}}}
\def\rvw{{\mathbf{w}}}
\def\rvx{{\mathbf{x}}}
\def\rvy{{\mathbf{y}}}
\def\rvz{{\mathbf{z}}}

\def\erva{{\textnormal{a}}}
\def\ervb{{\textnormal{b}}}
\def\ervc{{\textnormal{c}}}
\def\ervd{{\textnormal{d}}}
\def\erve{{\textnormal{e}}}
\def\ervf{{\textnormal{f}}}
\def\ervg{{\textnormal{g}}}
\def\ervh{{\textnormal{h}}}
\def\ervi{{\textnormal{i}}}
\def\ervj{{\textnormal{j}}}
\def\ervk{{\textnormal{k}}}
\def\ervl{{\textnormal{l}}}
\def\ervm{{\textnormal{m}}}
\def\ervn{{\textnormal{n}}}
\def\ervo{{\textnormal{o}}}
\def\ervp{{\textnormal{p}}}
\def\ervq{{\textnormal{q}}}
\def\ervr{{\textnormal{r}}}
\def\ervs{{\textnormal{s}}}
\def\ervt{{\textnormal{t}}}
\def\ervu{{\textnormal{u}}}
\def\ervv{{\textnormal{v}}}
\def\ervw{{\textnormal{w}}}
\def\ervx{{\textnormal{x}}}
\def\ervy{{\textnormal{y}}}
\def\ervz{{\textnormal{z}}}

\def\rmA{{\mathbf{A}}}
\def\rmB{{\mathbf{B}}}
\def\rmC{{\mathbf{C}}}
\def\rmD{{\mathbf{D}}}
\def\rmE{{\mathbf{E}}}
\def\rmF{{\mathbf{F}}}
\def\rmG{{\mathbf{G}}}
\def\rmH{{\mathbf{H}}}
\def\rmI{{\mathbf{I}}}
\def\rmJ{{\mathbf{J}}}
\def\rmK{{\mathbf{K}}}
\def\rmL{{\mathbf{L}}}
\def\rmM{{\mathbf{M}}}
\def\rmN{{\mathbf{N}}}
\def\rmO{{\mathbf{O}}}
\def\rmP{{\mathbf{P}}}
\def\rmQ{{\mathbf{Q}}}
\def\rmR{{\mathbf{R}}}
\def\rmS{{\mathbf{S}}}
\def\rmT{{\mathbf{T}}}
\def\rmU{{\mathbf{U}}}
\def\rmV{{\mathbf{V}}}
\def\rmW{{\mathbf{W}}}
\def\rmX{{\mathbf{X}}}
\def\rmY{{\mathbf{Y}}}
\def\rmZ{{\mathbf{Z}}}

\def\ermA{{\textnormal{A}}}
\def\ermB{{\textnormal{B}}}
\def\ermC{{\textnormal{C}}}
\def\ermD{{\textnormal{D}}}
\def\ermE{{\textnormal{E}}}
\def\ermF{{\textnormal{F}}}
\def\ermG{{\textnormal{G}}}
\def\ermH{{\textnormal{H}}}
\def\ermI{{\textnormal{I}}}
\def\ermJ{{\textnormal{J}}}
\def\ermK{{\textnormal{K}}}
\def\ermL{{\textnormal{L}}}
\def\ermM{{\textnormal{M}}}
\def\ermN{{\textnormal{N}}}
\def\ermO{{\textnormal{O}}}
\def\ermP{{\textnormal{P}}}
\def\ermQ{{\textnormal{Q}}}
\def\ermR{{\textnormal{R}}}
\def\ermS{{\textnormal{S}}}
\def\ermT{{\textnormal{T}}}
\def\ermU{{\textnormal{U}}}
\def\ermV{{\textnormal{V}}}
\def\ermW{{\textnormal{W}}}
\def\ermX{{\textnormal{X}}}
\def\ermY{{\textnormal{Y}}}
\def\ermZ{{\textnormal{Z}}}

\def\vzero{{\bm{0}}}
\def\vone{{\bm{1}}}
\def\vmu{{\bm{\mu}}}
\def\vtheta{{\bm{\theta}}}
\def\va{{\bm{a}}}
\def\vb{{\bm{b}}}
\def\vc{{\bm{c}}}
\def\vd{{\bm{d}}}
\def\ve{{\bm{e}}}
\def\vf{{\bm{f}}}
\def\vg{{\bm{g}}}
\def\vh{{\bm{h}}}
\def\vi{{\bm{i}}}
\def\vj{{\bm{j}}}
\def\vk{{\bm{k}}}
\def\vl{{\bm{l}}}
\def\vm{{\bm{m}}}
\def\vn{{\bm{n}}}
\def\vo{{\bm{o}}}
\def\vp{{\bm{p}}}
\def\vq{{\bm{q}}}
\def\vr{{\bm{r}}}
\def\vs{{\bm{s}}}
\def\vt{{\bm{t}}}
\def\vu{{\bm{u}}}
\def\vv{{\bm{v}}}
\def\vw{{\bm{w}}}
\def\vx{{\bm{x}}}
\def\vy{{\bm{y}}}
\def\vz{{\bm{z}}}

\def\evalpha{{\alpha}}
\def\evbeta{{\beta}}
\def\evepsilon{{\epsilon}}
\def\evlambda{{\lambda}}
\def\evomega{{\omega}}
\def\evmu{{\mu}}
\def\evpsi{{\psi}}
\def\evsigma{{\sigma}}
\def\evtheta{{\theta}}
\def\eva{{a}}
\def\evb{{b}}
\def\evc{{c}}
\def\evd{{d}}
\def\eve{{e}}
\def\evf{{f}}
\def\evg{{g}}
\def\evh{{h}}
\def\evi{{i}}
\def\evj{{j}}
\def\evk{{k}}
\def\evl{{l}}
\def\evm{{m}}
\def\evn{{n}}
\def\evo{{o}}
\def\evp{{p}}
\def\evq{{q}}
\def\evr{{r}}
\def\evs{{s}}
\def\evt{{t}}
\def\evu{{u}}
\def\evv{{v}}
\def\evw{{w}}
\def\evx{{x}}
\def\evy{{y}}
\def\evz{{z}}

\def\mA{{\bm{A}}}
\def\mB{{\bm{B}}}
\def\mC{{\bm{C}}}
\def\mD{{\bm{D}}}
\def\mE{{\bm{E}}}
\def\mF{{\bm{F}}}
\def\mG{{\bm{G}}}
\def\mH{{\bm{H}}}
\def\mI{{\bm{I}}}
\def\mJ{{\bm{J}}}
\def\mK{{\bm{K}}}
\def\mL{{\bm{L}}}
\def\mM{{\bm{M}}}
\def\mN{{\bm{N}}}
\def\mO{{\bm{O}}}
\def\mP{{\bm{P}}}
\def\mQ{{\bm{Q}}}
\def\mR{{\bm{R}}}
\def\mS{{\bm{S}}}
\def\mT{{\bm{T}}}
\def\mU{{\bm{U}}}
\def\mV{{\bm{V}}}
\def\mW{{\bm{W}}}
\def\mX{{\bm{X}}}
\def\mY{{\bm{Y}}}
\def\mZ{{\bm{Z}}}
\def\mBeta{{\bm{\beta}}}
\def\mPhi{{\bm{\Phi}}}
\def\mLambda{{\bm{\Lambda}}}
\def\mSigma{{\bm{\Sigma}}}

\newcommand{\tens}[1]{\bm{\mathsfit{#1}}}
\def\tA{{\tens{A}}}
\def\tB{{\tens{B}}}
\def\tC{{\tens{C}}}
\def\tD{{\tens{D}}}
\def\tE{{\tens{E}}}
\def\tF{{\tens{F}}}
\def\tG{{\tens{G}}}
\def\tH{{\tens{H}}}
\def\tI{{\tens{I}}}
\def\tJ{{\tens{J}}}
\def\tK{{\tens{K}}}
\def\tL{{\tens{L}}}
\def\tM{{\tens{M}}}
\def\tN{{\tens{N}}}
\def\tO{{\tens{O}}}
\def\tP{{\tens{P}}}
\def\tQ{{\tens{Q}}}
\def\tR{{\tens{R}}}
\def\tS{{\tens{S}}}
\def\tT{{\tens{T}}}
\def\tU{{\tens{U}}}
\def\tV{{\tens{V}}}
\def\tW{{\tens{W}}}
\def\tX{{\tens{X}}}
\def\tY{{\tens{Y}}}
\def\tZ{{\tens{Z}}}

\def\gA{{\mathcal{A}}}
\def\gB{{\mathcal{B}}}
\def\gC{{\mathcal{C}}}
\def\gD{{\mathcal{D}}}
\def\gE{{\mathcal{E}}}
\def\gF{{\mathcal{F}}}
\def\gG{{\mathcal{G}}}
\def\gH{{\mathcal{H}}}
\def\gI{{\mathcal{I}}}
\def\gJ{{\mathcal{J}}}
\def\gK{{\mathcal{K}}}
\def\gL{{\mathcal{L}}}
\def\gM{{\mathcal{M}}}
\def\gN{{\mathcal{N}}}
\def\gO{{\mathcal{O}}}
\def\gP{{\mathcal{P}}}
\def\gQ{{\mathcal{Q}}}
\def\gR{{\mathcal{R}}}
\def\gS{{\mathcal{S}}}
\def\gT{{\mathcal{T}}}
\def\gU{{\mathcal{U}}}
\def\gV{{\mathcal{V}}}
\def\gW{{\mathcal{W}}}
\def\gX{{\mathcal{X}}}
\def\gY{{\mathcal{Y}}}
\def\gZ{{\mathcal{Z}}}

\def\sA{{\mathbb{A}}}
\def\sB{{\mathbb{B}}}
\def\sC{{\mathbb{C}}}
\def\sD{{\mathbb{D}}}
\def\sF{{\mathbb{F}}}
\def\sG{{\mathbb{G}}}
\def\sH{{\mathbb{H}}}
\def\sI{{\mathbb{I}}}
\def\sJ{{\mathbb{J}}}
\def\sK{{\mathbb{K}}}
\def\sL{{\mathbb{L}}}
\def\sM{{\mathbb{M}}}
\def\sN{{\mathbb{N}}}
\def\sO{{\mathbb{O}}}
\def\sP{{\mathbb{P}}}
\def\sQ{{\mathbb{Q}}}
\def\sR{{\mathbb{R}}}
\def\sS{{\mathbb{S}}}
\def\sT{{\mathbb{T}}}
\def\sU{{\mathbb{U}}}
\def\sV{{\mathbb{V}}}
\def\sW{{\mathbb{W}}}
\def\sX{{\mathbb{X}}}
\def\sY{{\mathbb{Y}}}
\def\sZ{{\mathbb{Z}}}

\def\emLambda{{\Lambda}}
\def\emA{{A}}
\def\emB{{B}}
\def\emC{{C}}
\def\emD{{D}}
\def\emE{{E}}
\def\emF{{F}}
\def\emG{{G}}
\def\emH{{H}}
\def\emI{{I}}
\def\emJ{{J}}
\def\emK{{K}}
\def\emL{{L}}
\def\emM{{M}}
\def\emN{{N}}
\def\emO{{O}}
\def\emP{{P}}
\def\emQ{{Q}}
\def\emR{{R}}
\def\emS{{S}}
\def\emT{{T}}
\def\emU{{U}}
\def\emV{{V}}
\def\emW{{W}}
\def\emX{{X}}
\def\emY{{Y}}
\def\emZ{{Z}}
\def\emSigma{{\Sigma}}

\newcommand{\etens}[1]{\mathsfit{#1}}
\def\etLambda{{\etens{\Lambda}}}
\def\etA{{\etens{A}}}
\def\etB{{\etens{B}}}
\def\etC{{\etens{C}}}
\def\etD{{\etens{D}}}
\def\etE{{\etens{E}}}
\def\etF{{\etens{F}}}
\def\etG{{\etens{G}}}
\def\etH{{\etens{H}}}
\def\etI{{\etens{I}}}
\def\etJ{{\etens{J}}}
\def\etK{{\etens{K}}}
\def\etL{{\etens{L}}}
\def\etM{{\etens{M}}}
\def\etN{{\etens{N}}}
\def\etO{{\etens{O}}}
\def\etP{{\etens{P}}}
\def\etQ{{\etens{Q}}}
\def\etR{{\etens{R}}}
\def\etS{{\etens{S}}}
\def\etT{{\etens{T}}}
\def\etU{{\etens{U}}}
\def\etV{{\etens{V}}}
\def\etW{{\etens{W}}}
\def\etX{{\etens{X}}}
\def\etY{{\etens{Y}}}
\def\etZ{{\etens{Z}}}

\newcommand{\pdata}{p_{\rm{data}}}
\newcommand{\ptrain}{\hat{p}_{\rm{data}}}
\newcommand{\Ptrain}{\hat{P}_{\rm{data}}}
\newcommand{\pmodel}{p_{\rm{model}}}
\newcommand{\Pmodel}{P_{\rm{model}}}
\newcommand{\ptildemodel}{\tilde{p}_{\rm{model}}}
\newcommand{\pencode}{p_{\rm{encoder}}}
\newcommand{\pdecode}{p_{\rm{decoder}}}
\newcommand{\precons}{p_{\rm{reconstruct}}}

\newcommand{\laplace}{\mathrm{Laplace}} %

\newcommand{\Ls}{\mathcal{L}}
\newcommand{\emp}{\tilde{p}}
\newcommand{\lr}{\alpha}
\newcommand{\reg}{\lambda}
\newcommand{\rect}{\mathrm{rectifier}}
\newcommand{\softmax}{\mathrm{softmax}}
\newcommand{\sigmoid}{\sigma}
\newcommand{\softplus}{\zeta}
\newcommand{\KL}{D_{\mathrm{KL}}}
\newcommand{\Var}{\mathrm{Var}}
\newcommand{\standarderror}{\mathrm{SE}}
\newcommand{\Cov}{\mathrm{Cov}}
\newcommand{\normlzero}{L^0}
\newcommand{\normlone}{L^1}
\newcommand{\normltwo}{L^2}
\newcommand{\normlp}{L^p}
\newcommand{\normmax}{L^\infty}

\newcommand{\parents}{Pa} %

\let\ab\allowbreak

\maketitle

\begin{figure}[th]
    \centering
    \vspace{-8mm}
    \includegraphics[width=0.96\linewidth]{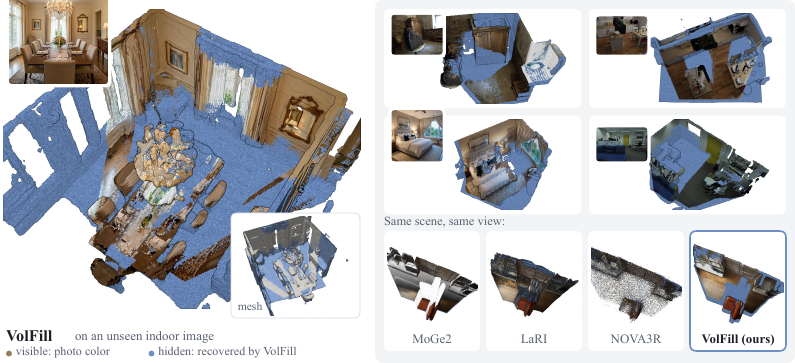}
    \vspace{1mm}
    \caption{\textbf{\Approach} recovers the \textit{complete} 3D scene, including surfaces hidden from the camera, from a single image. Geometry seen in the input view keeps the photo's color; hidden geometry recovered by \Approach is shown in \textcolor[HTML]{6E8FC4}{\textbf{blue}}. \textit{Left:} an unseen indoor image, as a point cloud and as a mesh extracted directly from the predicted TUDF grid. \textit{Top right:} more qualitative samples. \textit{Bottom right:} on the same scene and view, pixel-aligned MoGe2 leaves holes behind occluders, LaRI produces layered artifacts, and NOVA3R yields noisy, disintegrated points, whereas \Approach produces sharp and complete geometry.}
    \label{fig:teaser}
\end{figure}

\begin{abstract}

    Reconstructing the complete geometry of a scene from a single RGB image remains challenging---especially when inferring hidden structures where visual evidence is incomplete. We introduce \Approach, a generative framework that predicts the 3D structure of the complete scene rather than relying on traditional pixel-aligned regression. Our method utilizes a hybrid 3D VAE to compress sparse \emph{truncated unsigned distance function} grids  into a compact latent space, paired with a latent Diffusion Transformer that denoises this representation to recover the complete scene. We condition the generation on a geometry foundation model, leveraging rich spatial priors for robust reasoning. Unlike existing methods limited by per-ray constraints or unstructured point-cloud queries, \Approach provides a structured representation that supports direct surface extraction and occupancy queries at scale. Extensive experiments on the SCRREAM and NRGB-D datasets demonstrate that our approach significantly outperforms current amodal baselines, providing a robust foundation for holistic spatial understanding.

\end{abstract}

\section{Introduction}
\label{sec:intro}

This paper addresses amodal 3D scene reconstruction, seeking to recover the complete geometry, including observed and occluded structures, from a single RGB image. Beyond being inherently ill-posed, amodal reconstruction must infer hidden structures where geometric evidence is partially absent. This requires synthesizing under-determined regions that remain physically plausible and consistent with the observed scene. Such spatial awareness is critical for practical applications, as navigating or interacting with an environment requires a comprehensive understanding of the scene that extends well beyond the immediate line of sight.

Recent feed-forward methods~\cite{moge,moge2,depthpro,pixel_perfect_depth,dust3r,vggt,depthanything,mapanything,pi3,ecosplat,airsplat} can recover accurate 3D structure from images in seconds, but all share a hard constraint: \emph{pixel alignment}. Every predicted point lies on a source camera ray, limiting reconstruction strictly to visible surfaces and producing duplicated geometry in overlapping views. Optimization-based scene representations~\cite{Lombardi2019,Niemeyer2019DifferentiableVR,sitzmann2020implicit,mildenhall2021nerf,kerbl3Dgaussians} reconstruct complete geometry but require dense captures and per-scene optimization, which can take hours to compute. Large reconstruction models~\cite{lrm,Zhang2024GSLRMLR} and generative 3D models~\cite{Xiang2024Structured3L,Xiang2025NativeAC,Li2025TripoSGH3,yang2024hunyuan3d} demonstrate impressive 3D reconstruction quality from sparse inputs, but are predominantly optimized for object-centric scenarios, limiting their applicability to complex scenes with unbounded extents and challenging amodal structures.

At the scene level, LaRI~\cite{lari} proposes a \emph{layered ray intersection} representation, regressing multiple ordered surface intersections per camera ray. While this enables amodal recovery within a pixel-aligned framework, it remains a partial solution; the fixed layer count under-represents densely occluded structures while wasting capacity in open regions. Furthermore, it frequently yields layered artifacts (the ``multiple wall'' effect) rather than accurately recovering the actual unobserved surfaces. Alternatively, NOVA3R~\cite{nova3r} employs flow matching to directly generate \emph{amodal 3D point clouds}, but often yields noisy and disintegrated outputs (Fig.~\ref{fig:teaser}).

To address these limitations, we propose \Approach, a latent diffusion framework that predicts a complete Truncated Unsigned Distance Function (TUDF) voxel grid from a single image. A TUDF encodes the distance to the nearest scene surface, including occluded regions, as a continuous scalar field, enabling direct surface extraction via isosurfacing~\cite{meshudf} without requiring post-processing reconstruction of point clouds. Unlike point clouds, it scales well with scene complexity; unlike layered ray representations, it places no constraint on the number of recoverable occluded layers. \Approach consists of a hybrid 3D VAE that encodes the sparse TUDF grid into a compact dense latent space, and a Diffusion Transformer (DiT) trained with flow matching. To make amodal reasoning robust despite limited 3D training data, we introduce a dual-conditioning strategy that combines high-level image tokens with explicit visible geometry from a frozen MoGe2~\cite{moge2} model, grounding the generation in the observed scene while leveraging the strong geometric priors of this foundation model.

Trained on 3D-FRONT~\cite{3dfront} and ScanNet++~\cite{yeshwanth2023scannet++}, \Approach achieves state-of-the-art amodal reconstruction on the SCRREAM~\cite{scrream} and NRGB-D~\cite{nrgbd} benchmarks, synthesizing high-fidelity amodal geometry with significantly greater sharpness and structural accuracy than existing amodal methods (Fig.~\ref{fig:teaser}). In summary, our main contributions are:
\begin{itemize}[leftmargin=1.5em, itemsep=0pt, parsep=0pt, topsep=0pt]
\item \textbf{\Approach}, a generative framework that utilizes volumetric flow matching to recover complete scene-level geometry from single-view images.
\item \textbf{A hybrid 3D VAE} enabling efficient spatial compression of high-resolution TUDF grids to a compact latent space, facilitating high-fidelity reconstruction of complex amodal structures.
\item \textbf{A dual-conditioning strategy} leveraging geometry foundation models to integrate high-level image tokens with explicit visible geometry.
\end{itemize}

\section{Related Work}
\label{sec:related_work}
\myheading{Volumetric 3D representations} provide a flexible foundation for 3D vision, spanning applications from scene understanding to generative modeling. To mitigate the cubic complexity of dense grids, sparse convolutional engines~\cite{Graham2017SubmanifoldSC, minkowski, torchsparse} have become the standard for efficiently processing geometry along active surfaces. In scene-level reconstruction, ``lifting'' paradigms map 2D features into these volumes through depth-guided projection~\cite{sscnet}, ray sampling~\cite{monoscene}, or transformer-based voxel queries~\cite{voxformer, surroundocc}. Within the generative domain, early models applied diffusion directly to raw voxels~\cite{diffrf}, whereas contemporary frameworks leverage hierarchical latent diffusion~\cite{sdfusion, xcube} or rectified flow on sparse latents~\cite{Xiang2024Structured3L} to achieve high-fidelity structural synthesis. Building on these advances, our approach adopts a TUDF representation within a structured 3D grid, using sparse convolutions to capture complex scene geometry while maintaining high computational efficiency.

\myheading{Pixel-aligned single-view 3D reconstruction} recovers \emph{visible} surface geometry, evolving from early handcrafted features~\cite{hoiem2007recovering,saxena2005learning,6787109,saxena2008make3d} to deep architectures~\cite{wang2018learning,fu2018deep,bhat2021adabins,bts,eigen2014depth}. Progress in scaling has yielded models with remarkable zero-shot generalization~\cite{ranftl2020towards,zoedepth,metric3d,metric3dv2,unidepth,Guizilini2023TowardsZS,depthanything,depthanythingv2,depthpro}. Recently, methods have further refined geometric accuracy by jointly predicting intrinsics~\cite{unidepth, depthpro}, regressing dense 3D pointmaps~\cite{moge, moge2}, or utilizing diffusion priors~\cite{stable_diffusion,stable_diffusion_xl} for high-fidelity, arbitrary-resolution depth synthesis~\cite{marigold, lotus,geowizard,finetune_diffusion,sharpdepth,pixel_perfect_depth,infinidepth}. Nonetheless, these approaches remain fundamentally confined to 2.5D visible surfaces. Their inability to infer occluded regions or amodal structure directly motivates our shift from surface-level estimation to full-scene reconstruction.

\myheading{Geometry foundation models as priors.} Recent geometry foundation models (GFMs)~\cite{dust3r, dens3r, moge, moge2, vggt, depthanything3} have demonstrated strong performance in 3D reconstruction and serve as robust feature extractors for downstream tasks requiring explicit spatial priors~\cite{pixel_perfect_depth, dage, geometryforcing, reconviagen}. Current approaches typically adopt one of two strategies: (1) using explicit geometric outputs (e.g., depth or point clouds) as structural anchors to guide synthesis~\cite{viewcrafter, trajectorycrafter, cognvs, occany}, or (2) injecting latent foundation features into transformer blocks to maintain consistency~\cite{geometryforcing, gld, lavr, taro}. Our approach unifies these paradigms via a dual-conditioning strategy. By leveraging both explicit visible geometry for structural grounding and foundation latent tokens for high-level context, we enable robust and physically plausible amodal reconstruction that transfers to held-out real indoor benchmarks.

\myheading{Amodal 3D reconstruction} aims to recover complete geometry from partial visual observations. Object-centric strategies often utilize two-stage pipelines---generating novel views via multi-view diffusion~\cite{zero123,syncdream,mvdream, wonder3d} before reconstructing geometry~\cite{lrm,instant3d,gtr,lgm,grm}---or synthesize 3D representations directly~\cite{point-e, single-stage-diff-nerf, Li2025TripoSGH3, yang2024hunyuan3d, Xiang2024Structured3L}, yet remain limited to isolated assets. For scenes, recent methods leverage camera trajectories for view synthesis~\cite{cameractrl, gen3c, viewcrafter, cat3d, seva, 4realvideo} but typically lack direct 3D outputs. Most related are LaRI~\cite{lari}, using layered ray intersections, and NOVA3R~\cite{nova3r}, which generates point clouds from scene tokens. \textbf{Concurrent with our work}, World Tracing~\cite{worldtracing} generates layered pixel-aligned geometry beyond the visible surface, while Surflo~\cite{surflo} and GenRecon~\cite{genrecon} reconstruct scenes from multiple views. In contrast, our structured volumetric representation enables more complete and accurate amodal reconstruction, recovering geometry with smoother surfaces and sharper structural detail than unstructured point-based approaches.

\section{Method}
\label{sec:method}
\subsection{Problem Formulation}
\label{sec:problem_formulation}

We address the task of amodal scene reconstruction from a single RGB image. Unlike traditional pixel-aligned methods~\cite{dust3r,depthanything, depthpro,moge,moge2} that recover only visible surfaces (e.g., \textit{depth maps} or \textit{pointmaps}), our goal is to predict the complete 3D geometry within the observed view frustum. This includes recovering amodal geometry, surfaces occluded by foreground objects, while maintaining the structural consistency of the scene-level environment.

\myheading{Input and output.} Given a single input image $I \in \mathbb{R}^{H \times W \times 3}$, our model predicts a complete volumetric grid $V \in \mathbb{R}^{N \times N \times N}$ at a resolution of $N=256$. We represent the scene geometry using a \emph{Truncated Unsigned Distance Function}. Let $\mathcal{S} \subset \mathbb{R}^3$ denote the set of all physical surfaces in the scene contained within the view frustum. For each voxel center $p$ within our grid, the value $V(p)$ is defined as the distance to the nearest surface in $\mathcal{S}$, truncated to a maximum value $\tau$:
{
\abovedisplayskip=6pt
\belowdisplayskip=3pt
\begin{equation}
    V(p) = \min(\text{dist}(p, \mathcal{S}), \tau)
\end{equation}
}

We employ an unsigned formulation because indoor scenes are frequently non-watertight. A signed distance function requires a globally consistent inside/outside orientation---well-defined only for closed, watertight surfaces. For the common open geometries (single-sided walls, floors, furniture), no such orientation exists, making the sign geometrically meaningless beyond the immediate voxel neighborhood. The unsigned formulation avoids this ambiguity, as distance-to-nearest-surface is well-defined for any geometry regardless of topology.

\myheading{Amodal geometry voxelization.} The amodal surfaces $\mathcal{S}$ can be obtained either directly from ground-truth dataset 3D meshes (if available) or by fusing depth from multiple surrounding views and transforming it into the target camera coordinate frame. To translate these amodal surfaces into our discrete target grid $V$, we employ a dynamic, frustum-aligned discretization strategy:
\begin{itemize}[leftmargin=1.5em, itemsep=0pt, parsep=1pt, topsep=1pt]
\item \emph{Spatial scope:} To establish the precise boundaries of our volumetric grid, we compute an axis-aligned bounding box $\mathcal{B}$ based on the \emph{visible} geometry $P_{\text{vis}} \subset \mathcal{S}$. Any amodal surfaces in $\mathcal{S}$ extending beyond $\mathcal{B}$ are discarded, focusing the representation on the immediate captured scene.
\item \emph{Dynamic scaling and TUDF computation:} Rather than using fixed spatial extents~\cite{monoscene,voxformer,surroundocc}, we calculate a dynamic, isotropic voxel size derived from the maximum dimension of $\mathcal{B}$ and our target resolution $N$. We then discretize this bounded space and compute the truncated distance from each voxel center $p$ to the cropped amodal surfaces, yielding the final TUDF grid $V$.
\end{itemize}

\myheading{Downstream geometric extractions.} The TUDF grid $V$ serves as a flexible geometric proxy supporting diverse downstream formats. Surfaces can be efficiently extracted as point clouds by thresholding voxels with distance values below a small threshold $V(p) < \epsilon_{\text{tudf}}$, or converted into meshes via MeshUDF~\cite{meshudf}, producing clean, topologically consistent reconstructions.

\begin{figure}[t]
    \centering
    \includegraphics[width=\linewidth]{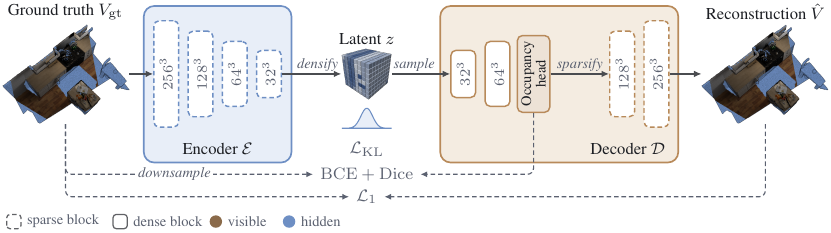}
    \vspace{-3mm}
    \caption{\textbf{3D VAE architecture.} The encoder compresses high-resolution sparse TUDF grids into a regularized dense latent via sparse convolutions. The decoder upsamples through dense layers, applies occupancy-guided sparsification, then restores the full-resolution TUDF via sparse convolutions.}
    \label{fig:3dvae}
    \vspace{-4mm}
\end{figure}

\subsection{Amodal 3D Reconstruction via a Generative Framework}

We formulate amodal scene reconstruction as estimating the conditional distribution $P(V|I)$ rather than a deterministic point estimate, overcoming a critical limitation of regression-based pipelines~\cite{moge, depthpro}. Standard regression losses (e.g., $L_2$) predict the conditional expectation $\mathbb{E}[V|I]$, averaging over plausible geometries and producing over-smoothed reconstructions in occluded regions where evidence is absent. In contrast, our generative framework samples from $P(V|I)$, producing sharp and physically plausible completions by following the learned manifold of real-world scene structures.

\subsection{Hybrid 3D VAE}
\label{sec:vae}

To bridge the gap between high-resolution TUDF grids and the manageable latent space required by our generative model, we propose a hybrid sparse-dense 3D VAE (Fig.~\ref{fig:3dvae}). This architecture exploits the inherent sparsity of scene geometry, where typically only 3--5\% of voxels are surface-adjacent, while providing a fixed-size dense latent representation suitable for denoising via a latent transformer.

\myheading{The encoder $\mathcal{E}$} is designed to process natively sparse TUDF inputs with high computational efficiency. We convert the input grids into sparse tensors, retaining only active voxels within the truncation band $\tau$, and apply a sequence of sparse 3D convolutions~\cite{spconv,torchsparse,minkowski} with strided downsampling to reduce the spatial resolution by $16\times$. At the bottleneck, these features are densified into a regular 4D tensor $z \in \mathbb{R}^{16 \times 16 \times 16 \times C}$; by densifying only at this highly compressed scale, we maintain a manageable memory footprint while ensuring compatibility with dense generative architectures. Finally, a standard KL-divergence bottleneck regularizes the latent space to facilitate stable sampling~\cite{stable_diffusion}.

\myheading{The decoder $\mathcal{D}$} faces the challenge of recovering a $256^3$ grid from a dense $16^3$ latent without prior knowledge of the target scene's sparsity pattern. We address this through a hybrid dense-to-sparse decoding schedule: 
\begin{itemize}[leftmargin=1.5em, itemsep=0pt, parsep=1pt, topsep=1pt]
    \item \emph{Dense upsampling} ($16^3\to 64^3$): We first employ standard transposed 3D convolutions to upsample the latent to a $64^3$ dense feature map, where the total voxel count ($\sim262\text{K}$) remains tractable for dense computation.
    
    \item \emph{Structure prediction}: At the $64^3$ stage, a binary occupancy head predicts a mask $\hat{O}$, indicating surface-adjacent voxels. This mask is used to sparsify the dense feature map, discarding empty regions and focusing subsequent computation solely on relevant geometric structures. We supervise this with a ground-truth occupancy mask $O_{\text{gt}}$ derived from the TUDF grid.
    
    \item \emph{Sparse upsampling} ($64^3 \to 256^3$): The sparse features are processed by sparse 3D convolutions to reach the final $256^3$ resolution, where memory cost scales linearly with surface area rather than cubically with volume, enabling high-resolution TUDF reconstruction at surface-adjacent voxels. The resulting sparse predictions are finally scattered back into a dense grid to obtain $\hat{V}$.
\end{itemize}

The VAE is trained end-to-end on our collected TUDF datasets using a composite loss function:
{
\abovedisplayskip=4pt
\belowdisplayskip=4pt
\begin{equation}
\label{eqn:vae_loss}
    \mathcal{L}_{\text{VAE}} = \mathcal{L}_1(\hat{V}, V_{\text{gt}}) + \lambda_{\text{bce}} \text{BCE}(\hat{O}, O_{\text{gt}}) + \lambda_{\text{dice}} \text{Dice}(\hat{O}, O_{\text{gt}}) + \lambda_{\text{kl}} \mathcal{L}_{\text{KL}}
\end{equation}
}
where $\mathcal{L}_1$ ensures precise distance reconstruction in active voxels, BCE and Dice losses supervise binary occupancy in the dense-to-sparse transition, and $\mathcal{L}_{\text{KL}}$ regularizes the latent distribution. This hybrid design facilitates aggressive $16\times$ spatial compression while preserving the fine-grained details essential for high-fidelity amodal reconstruction.

\begin{figure}[t]
    \centering
    \includegraphics[width=\linewidth]{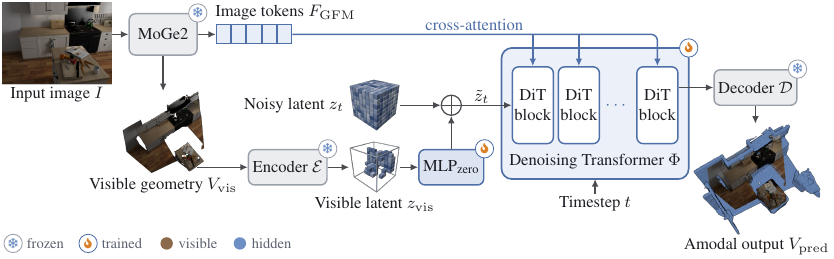}
    \vspace{-4.5mm}
    \caption{\textbf{Latent DiT architecture.} It operates in the compressed VAE latent space using a denoising transformer backbone with a flow-matching objective. We leverage a \textbf{dual-conditioning} strategy, integrating high-level image tokens and explicit visible geometry, to guide the generative process and synthesize sharp, scene-consistent amodal structures.}
    \label{fig:dit}
    \vspace{-3mm}
\end{figure}

\subsection{Geometry-Conditioned Flow Matching}

We implement a latent Diffusion Transformer $\Phi$ that learns the conditional velocity field to transport noise toward the distribution of scene TUDF latents. Starting from $z_{0} \sim \mathcal{N}(0, \mathbf{I})$, we iteratively integrate the learned flow to obtain a clean latent $z_1$, which is decoded by $\mathcal{D}$ to produce the final TUDF prediction $V_{\text{pred}}$. To guide this process, we propose a dual-conditioning strategy that supplements global semantic-geometric context with local structural anchors. This architecture is shown in Fig.~\ref{fig:dit}.

\myheading{Global geometric priors.} To inherit a robust, zero-shot understanding of spatial layouts, we condition the DiT on frozen features ($F_{\text{GFM}}$) extracted from a geometry foundation model. We utilize MoGe2~\cite{moge2} as our backbone for its strong geometric estimation performance. Following standard practice, these features are processed via layer normalization and a linear projection before being injected into the cross-attention layers of each DiT block as keys and values (details in Appendix~\ref{sec:supp_arch}).

\myheading{Visible-latent inpainting.} While global features provide context, we explicitly anchor the generative process to the observed scene by integrating known visible geometry directly into the latent space. Conceptually motivated by unprojection-based synthesis~\cite{viewcrafter, cognvs}, this strategy provides a partial guide for the model to complete occluded regions. Concretely, we obtain the visible pointmap $P_{\text{vis}}$ from the GFM and convert it into a TUDF grid. This is then encoded into a visible latent $z_{\text{vis}}$ via the frozen VAE encoder $\mathcal{E}$, and fused with the noisy latent as:
{
\abovedisplayskip=3pt
\belowdisplayskip=3pt
\begin{equation}
    \tilde{z}_t = z_t + \text{MLP}_{\text{zero}}(z_{\text{vis}})
\end{equation}
}where $\text{MLP}_{\text{zero}}$ is a zero-initialized projection layer. This provides the model with an explicit structural prior over observed regions to guide completion of occluded geometry.

\begin{table}[t]
\centering
\small
\caption{\textbf{3D reconstruction results} on the SCRREAM dataset. We mark {\setlength{\fboxsep}{2pt}\colorbox{custom_green!70}{\strut best}} and {\setlength{\fboxsep}{2pt}\colorbox{custom_green!30}{\strut second-best}}.}
\setlength{\tabcolsep}{3pt}
\small
\renewcommand{\arraystretch}{1.15}
\begin{threeparttable}
\begin{tabular}{lccccccccc}
\toprule
\multirow{2}{*}{\textbf{Method}}  & \multicolumn{3}{c}{\textcolor{black!40}{\textbf{Visible}}} & \multicolumn{3}{c}{\textbf{Occluded}} & \multicolumn{3}{c}{\textbf{Complete}} \\
\cmidrule(lr){2-4} \cmidrule(lr){5-7} \cmidrule(lr){8-10}
& \textcolor{black!40}{CD$\downarrow$} & \textcolor{black!40}{APD$_{0.1}\uparrow$} & \textcolor{black!40}{APD$_{0.02}\uparrow$} & CD$\downarrow$ & APD$_{0.1}\uparrow$ & APD$_{0.02}\uparrow$ & CD$\downarrow$ & FS$_{0.1}\uparrow$ & FS$_{0.02}\uparrow$ \\
\midrule
TripoSG$^\dagger$~\cite{Li2025TripoSGH3} & \color{black!40}13.75 & \color{black!40}44.74 & \color{black!40}10.08 & 10.73 & 55.34 & 13.37 & 11.10 & 53.56 & 13.30 \\
TRELLIS$^\dagger$~\cite{Xiang2024Structured3L} & \color{black!40}11.63 & \color{black!40}54.05 & \color{black!40}15.00 & 9.65 & 60.70 & 17.52 & 10.72 & 56.69 & 15.59 \\
\midrule
DUSt3R~\cite{dust3r} & \color{black!40}4.46 & \color{black!40}92.13 & \color{black!40}30.81 & 23.21 & 38.43 & 8.40 & 10.85 & 69.90 & 24.17 \\
VGGT~\cite{vggt} & \color{black!40}3.04 & \color{black!40}96.98 & \color{black!40}45.62 & 17.40 & 44.60 & 10.04 & 6.77 & 80.45 & 36.77 \\
DA3~\cite{depthanything3} & \cellcolor{custom_green!30}\color{black!40}2.37 & \cellcolor{custom_green!30}\color{black!40}98.98 & \color{black!40}54.34 & 16.16 & 48.31 & 10.76 & 6.16 & 82.86 & 43.10 \\
\cdashline{1-10}
DepthPro~\cite{depthpro} & \color{black!40}3.85 & \color{black!40}95.10 & \color{black!40}33.30 & 15.31 & 48.66 & 9.71 & 6.90 & 80.36 & 28.44 \\
MoGe2~\cite{moge2} &  \cellcolor{custom_green!70}\color{black!40}2.28 & \cellcolor{custom_green!70}\color{black!40}99.38 & \cellcolor{custom_green!70}\color{black!40}57.42 & 15.29 & 50.09 & 10.30 & 5.74 & 83.96 & \cellcolor{custom_green!30}45.84 \\
\cdashline{1-10}
LaRI~\cite{lari} & \color{black!40}3.43 & \color{black!40}95.48 & \color{black!40}41.61 & 5.23 & 85.25 & 29.31 & 5.19 & 85.43 & 30.85 \\
NOVA3R~\cite{nova3r} & \color{black!40}3.20 & \color{black!40}96.77 & \color{black!40}43.75 & \cellcolor{custom_green!30}3.56 & \cellcolor{custom_green!70}94.86 & \cellcolor{custom_green!30}41.21 & \cellcolor{custom_green!30}3.43 & \cellcolor{custom_green!70}95.26 & 45.12 \\
\cdashline{1-10}
\textbf{\Approach (ours)} & \color{black!40}2.84 & \color{black!40}96.10 & \cellcolor{custom_green!30}\color{black!40}56.73 & \cellcolor{custom_green!70}3.46 & \cellcolor{custom_green!30}92.70 & \cellcolor{custom_green!70}55.53 & \cellcolor{custom_green!70}3.03 & \cellcolor{custom_green!30}95.03 & \cellcolor{custom_green!70}54.83 \\
\bottomrule
\end{tabular}
\begin{tablenotes}\footnotesize
\item \emph{Note:} All reported numbers are scaled by a factor of $10^2$. \quad $^{\dagger}$: methods trained on object-centric datasets. \end{tablenotes}
\end{threeparttable}
\label{tab:scrream_results}
\vspace{-5mm}
\end{table}

\myheading{Flow matching objective.} We train our model using a rectified flow objective~\cite{Lipman2022FlowMF,Liu2022FlowSA,albergo2023building}. Given a ground-truth TUDF latent $z_1$ and Gaussian noise $\epsilon \sim \mathcal{N}(0, \mathbf{I})$, we define a linear interpolation path $z_t = (1-t)\epsilon + tz_1$. Our training objective is to minimize the expected mean squared error between the predicted velocity $u_\Phi$ and the constant velocity target $u^* = z_1 - \epsilon$:
{
\abovedisplayskip=4pt
\belowdisplayskip=4pt
\begin{equation}
   \mathcal{L}_{\text{gen}} =  \mathbb{E}_{t, \epsilon, z_1} \left[\ \|u_{\Phi}(\tilde{z}_t, t, F_{\text{GFM}}) - u^*\|^2 \right]
\end{equation}
}
During inference, we integrate the learned ODE from $t=0$ to $t=1$ via Euler steps, then decode the resulting latent through the VAE decoder $\mathcal{D}$ to produce the final TUDF grid $V_{\text{pred}}$.

\subsection{Model Architecture}
\myheading{The 3D VAE} employs a symmetric encoder-decoder architecture, where each stage consists of two ResNet-style blocks and doubles the channel dimension at each resolution step. The primary distinction lies in the convolution implementation: while the encoder uses sparse 3D convolutions across all stages for efficiency, the decoder utilizes standard dense convolutions at lower resolutions ($16^3 \to 64^3$) before transitioning to sparse convolutions for the remaining higher stages ($64^3 \to 256^3$). Finally, the binary occupancy and TUDF regression heads are both implemented as 2-layer MLPs.

\myheading{The latent DiT} features 12 transformer blocks with a 768-dimensional hidden state and 12-head self-attention~\cite{attention}. Each block integrates image tokens $F_{\text{GFM}}$ through cross-attention, utilizes AdaLN~\cite{dit} for timestep conditioning, and employs a GELU feed-forward network~\cite{gelu} with $4\times$ expansion. To stabilize flow-matching training, we apply QK-normalization~\cite{qknorm} via RMSNorm~\cite{rmsnorm} before the attention layers. The detailed architecture of the DiT block is provided in Appendix~\ref{sec:supp_arch}.

\section{Experiments}
\label{sec:exp}

\subsection{Implementation Details}

\myheading{Training datasets.} We train \Approach on a combination of 3D-FRONT~\cite{3dfront} and ScanNet++~\cite{yeshwanth2023scannet++}, following the data splits established by LaRI~\cite{lari} and NOVA3R~\cite{nova3r}. For the synthetic 3D-FRONT dataset, we utilize 18k room-level scenes to construct 96k image-TUDF pairs. For real-world ScanNet++ data, we aggregate multi-view depth maps and back-project them into a unified point cloud, followed by a filtering and voxelization pipeline to obtain amodal TUDF grids for 46k samples. All TUDF grids use $N=256$ and truncation $\tau=3$ voxels within a frustum-aligned box estimated from the visible geometry with robust percentile clipping and 10\% padding.

\myheading{Training protocol.} Training is conducted on two NVIDIA A6000/L40S GPUs in two stages: in Stage 1, the 3D VAE is trained for 20 epochs with a total batch size of 24, requiring 3 days to converge; in Stage 2, the latent transformer is trained for 100 epochs with a total batch size of 48, completing in 2.5 days. Both stages utilize the AdamW optimizer with an initial learning rate of $10^{-4}$, a cosine scheduler, and mixed-precision training. More training details are provided in Appendix~\ref{sec:supp_imp_details}.

\myheading{Latent sampling and inference.} During training, we sample the flow-matching timestep $t \in [0,1]$ from a logit-normal distribution to prioritize training on higher-noise regimes following~\cite{Xiang2024Structured3L}. 
We apply classifier-free guidance (CFG) by dropping conditions with a probability of $0.1$. At inference, we utilize an ODE solver with 50 steps and set the CFG scale to $3.0$. On a single RTX 4090, inference takes 1.4\,s per image and peaks at about 3.0\,GB of GPU memory for generation and 3.5\,GB end to end, including VAE decoding. Inference code and pretrained models are publicly available; training, evaluation, and data-processing code, together with the processed TUDF data and split files, will also be released.

\begin{table}[t]
\centering
\small
\caption{\textbf{3D reconstruction results} on the NRGB-D dataset ($\times 10^2$).}
\setlength{\tabcolsep}{3.5pt}
\small
\renewcommand{\arraystretch}{1.15}
\begin{tabular}{lcccccccccccc}
\toprule
\multirow{2}{*}{\textbf{Method}}  & \multicolumn{3}{c}{\textcolor{black!40}{\textbf{Visible}}} & \multicolumn{3}{c}{\textbf{Occluded}} & \multicolumn{3}{c}{\textbf{Complete}}  \\
\cmidrule(lr){2-4} \cmidrule(lr){5-7} \cmidrule(lr){8-10}
& \textcolor{black!40}{CD$\downarrow$} & \textcolor{black!40}{APD$_{0.1}\uparrow$} & \textcolor{black!40}{APD$_{0.05}\uparrow$} & CD$\downarrow$ & APD$_{0.1}\uparrow$ & APD$_{0.05}\uparrow$ & CD$\downarrow$ & FS$_{0.1}\uparrow$ & FS$_{0.05}\uparrow$ \\
\midrule
DUSt3R~\cite{dust3r} & \color{black!40}4.46 & \color{black!40}92.13 & \color{black!40}69.08 & 23.21 & 38.43 & 22.18 & 10.85 & 69.90 & 51.61 \\
VGGT~\cite{vggt} & \color{black!40}3.25 & \color{black!40}96.51 & \color{black!40}81.63 & 21.38 & 42.15 & 26.35 & 9.44 & 73.85 & 58.91 \\
DA3~\cite{depthanything3} & \cellcolor{custom_green!30}\color{black!40}2.69 & \cellcolor{custom_green!70}\color{black!40}98.68 & \cellcolor{custom_green!30}\color{black!40}89.24 & 21.65 & 43.34 & 28.15 & 8.88 & 74.87 & \cellcolor{custom_green!30}63.52 \\
\cdashline{1-10}
DepthPro~\cite{depthpro} & \color{black!40}4.38 & \color{black!40}92.56 & \color{black!40}67.41 & 22.19 & 41.69 & 24.64 & 10.03 & 71.30 & 51.60 \\
MoGe2~\cite{moge2} & \cellcolor{custom_green!70}\color{black!40}2.62 & \cellcolor{custom_green!30}\color{black!40}98.46 & \cellcolor{custom_green!70}\color{black!40}89.55 & 21.12 & 43.31 & 28.43 & 9.31 & \cellcolor{custom_green!30}75.35 & 61.93 \\
\cdashline{1-10}
LaRI~\cite{lari} & \color{black!40}4.09 & \color{black!40}93.77 & \color{black!40}73.51 & 13.04 & 64.53 & 45.82 & 9.79 & 68.37 & 43.47 \\
NOVA3R~\cite{nova3r} & \color{black!40}5.05 & \color{black!40}89.63 & \color{black!40}64.44 & \cellcolor{custom_green!30}8.31 & \cellcolor{custom_green!30}80.50 & \cellcolor{custom_green!30}58.55 & \cellcolor{custom_green!30}8.19 & 73.99 & 49.92 \\
\cdashline{1-10}
\textbf{\Approach (ours)} & \color{black!40}4.52 & \color{black!40}90.06 & \color{black!40}69.22 & \cellcolor{custom_green!70}6.97 & \cellcolor{custom_green!70}81.41 & \cellcolor{custom_green!70}62.48 & \cellcolor{custom_green!70}6.44 & \cellcolor{custom_green!70}84.35 & \cellcolor{custom_green!70}63.92  \\
\bottomrule
\end{tabular}
\label{tab:nrgbd_results}
\vspace{-5mm}
\end{table}

\subsection{Comparison with Prior Work}
\myheading{Datasets.} We evaluate our model on two benchmarks. (1) SCRREAM~\cite{scrream} consists of 460 samples with complete, high-quality scanned meshes, providing reliable ground truth for both visible and occluded surfaces. (2) Neural RGB-D (NRGB-D)~\cite{nrgbd} is adapted for amodal evaluation by aggregating its depth maps into a global coordinate frame via camera trajectories and multi-view fusion to reconstruct occluded surfaces. To ensure the benchmark tests scene completion, we retain only frames with substantial hidden geometry using an objective rule: we crop the fused geometry to each frame's view frustum and keep frames whose amodal-enrichment ratio (occluded to visible voxels) is at least $0.2$, yielding 137 samples from 9 scenes. We will release the curation script and the sample list.

\myheading{Metrics.} Following~\cite{lari,nova3r}, we evaluate reconstruction quality using Chamfer distance (CD$\downarrow$) and F-score (FS$_{\gamma}\uparrow$) at thresholds $\gamma \in \{0.1, 0.02\}$ ($\{0.1, 0.05\}$ on NRGB-D). Since our non-pixel-aligned formulation lacks point-wise correspondences, we recover the optimal similarity transformation (scale, rotation, translation) by minimizing Chamfer distance via gradient descent prior to evaluation; this 7-DoF alignment and the evaluation protocol of NOVA3R~\cite{nova3r} are applied identically to all methods. We assess performance across \emph{visible}, \emph{occluded}, and \emph{complete} subsets. We employ a one-way Chamfer distance (measuring from ground truth to prediction) in the \emph{visible} and \emph{occluded} subsets to assess regional coverage without penalizing valid predicted geometry that falls outside the target subset, and replace the F-score with a threshold coverage score APD$_{\gamma}\uparrow$, measuring the percentage of ground-truth points successfully reconstructed within threshold $\gamma$. We further evaluate generative quality via Fréchet Point Cloud Distance (FPD$\downarrow$)~\cite{seen2scene}, using a pretrained Uni3D~\cite{uni3d} to measure distributional similarity in a semantically aligned 3D feature space.

\myheading{Baselines.} We compare against four groups. \emph{Object-level generative models} (TRELLIS~\cite{Xiang2024Structured3L}, TripoSG~\cite{Li2025TripoSGH3}) show the effect of naively applying strong 3D object priors to full unmasked scenes. \emph{Multi-view geometry} models (DUSt3R~\cite{dust3r}, VGGT~\cite{vggt}, DepthAnything3~\cite{depthanything3}) and \emph{single-view pixel-aligned} estimators (MoGe2~\cite{moge2}, DepthPro~\cite{depthpro}) establish visible-surface reference performance but fundamentally cannot recover hidden geometry. Finally, \emph{scene-level amodal reconstruction} methods (LaRI~\cite{lari}, NOVA3R~\cite{nova3r}) serve as our primary competitors.

\begin{figure}[t]
    \centering
    \includegraphics[width=0.96\linewidth]{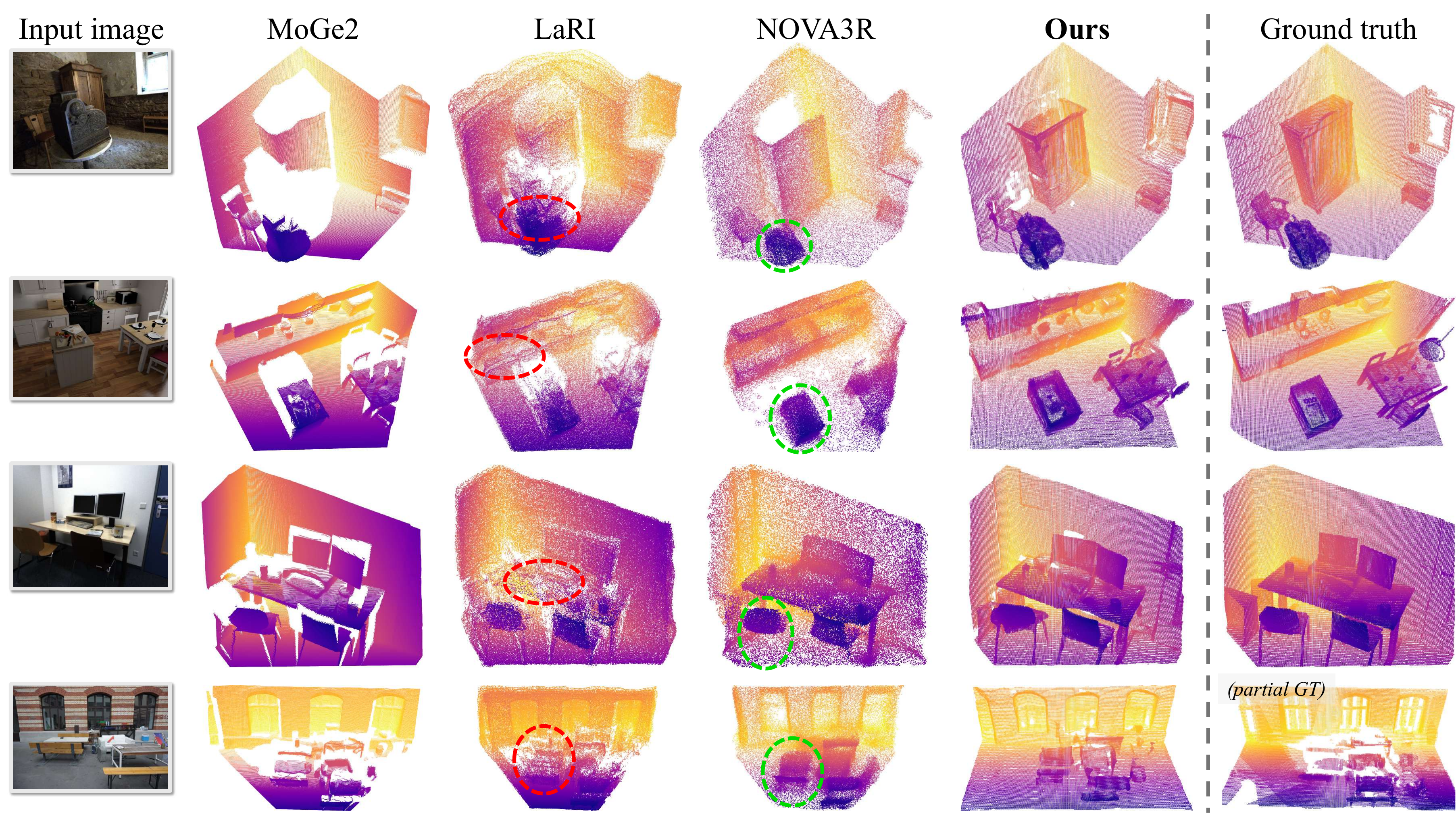}
    \vspace{-2mm}
    \caption{\textbf{Qualitative comparison.} \Approach synthesizes sharp, high-fidelity geometry, whereas LaRI produces layered artifacts (\textcolor[RGB]{255,0,0}{red circle}) and holes, and NOVA3R yields noisy, unstructured point scatters (\textcolor[RGB]{0,255,0}{green circle}).}
    \label{fig:quali_results}
    \vspace{-5mm}
\end{figure}

\begin{figure}[t]
    \centering
    \includegraphics[width=0.96\linewidth]{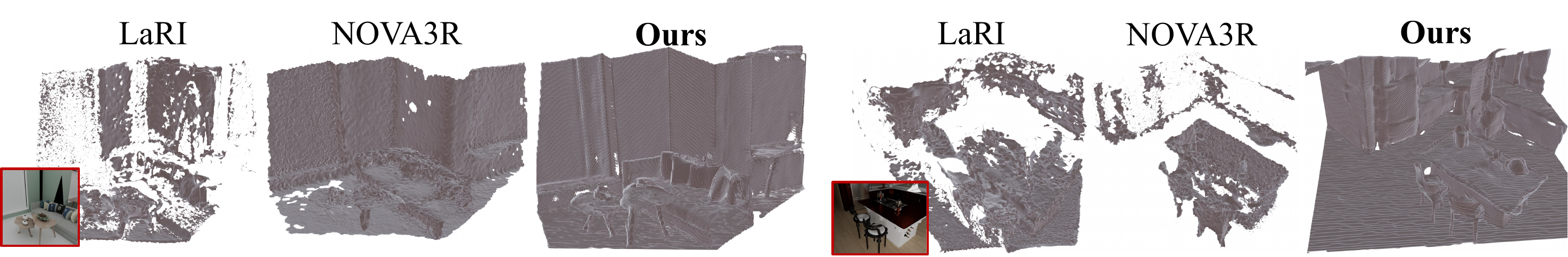}
    \vspace{-2mm}
    \caption{\textbf{Mesh reconstruction comparison.} LaRI and NOVA3R produce fragmented and noisy meshes due to their unstructured outputs, whereas \Approach directly extracts clean, topologically consistent surfaces from the structured TUDF grid.}
    \label{fig:quali_mesh_results}
    \vspace{-5mm}
\end{figure}

\myheading{Results.} Tables~\ref{tab:scrream_results} and~\ref{tab:nrgbd_results} report performance on SCRREAM and NRGB-D, respectively. While pixel-aligned estimators~\cite{moge2, depthanything3} naturally dominate visible metrics, they inherently fail to reconstruct occluded regions. Among methods designed for amodal completion~\cite{lari, nova3r}, our approach establishes a new state of the art for complete scene reconstruction, achieving a significant performance leap under the stringent $\text{FS}_{0.02}$ metric on SCRREAM. On NRGB-D visible metrics, we maintain performance near that of LaRI~\cite{lari}, while achieving substantially stronger reconstruction across occluded and complete geometry---confirming that our volumetric formulation provides a far more robust foundation for full-scene understanding than existing point-cloud or ray-based baselines.

\myheading{Is the representation the bottleneck?} Our visible accuracy is slightly below that of MoGe2 (Table~\ref{tab:scrream_results}). To locate this gap, we pass ground-truth geometry through our representation on all 460 SCRREAM samples and score the recovered visible surface against the ground truth: converting it into our $256^3$ TUDF grid and extracting the surface back yields a median visible $\text{APD}_{0.02}$ of $99.9$, and additionally encoding and decoding it with our VAE also yields $99.9$. As our full model reaches $66.3$ on the same measure, neither the grid resolution nor the VAE limits accuracy; the gap arises in the generative step, which re-synthesizes the visible surface from soft conditioning rather than copying it.

\myheading{Seed stability and diversity.} All tables report a single sample per image. Over five sampling seeds on SCRREAM, results are stable (complete CD $3.11\pm0.08$, $\text{FS}_{0.02}$ $54.2\pm0.5$), while the variation between samples is about $1.3\times$ larger in occluded than in visible regions, indicating that \Approach produces diverse completions of ambiguous hidden geometry while keeping the observed surface consistent.

Table~\ref{tab:metric_evaluation} presents our metric-scale evaluation, where we assess the reconstruction of amodal point clouds by applying a rigid transformation to align the predicted and ground-truth coordinates. While our model leverages internal MoGe2 predictions to infer absolute scale, we provide baselines~\cite{lari, nova3r} with a standalone MoGe2 model to lift their normalized outputs into metric space. Our approach significantly outperforms these methods, confirming its ability to reconstruct faithful metric-scale amodal geometry. Our superior FPD score demonstrates that the synthesized geometry is distributionally well-aligned with ground-truth scenes within a semantically meaningful feature space.

Fig.~\ref{fig:quali_results} presents qualitative comparisons of predicted point clouds. LaRI produces severe layered artifacts inherited from its ray-aligned formulation and leaves many occluded regions empty; NOVA3R recovers broader scene coverage but produces a noisy, unstructured point cloud. \Approach recovers more complete geometry with higher structural fidelity, especially in occluded regions. 

Furthermore, the structured nature of our TUDF representation enables direct surface extraction~\cite{meshudf}, yielding smooth and topologically consistent meshes, as qualitatively evidenced in Fig.~\ref{fig:quali_mesh_results}. In contrast,~\cite{lari, nova3r} require Poisson reconstruction~\cite{poissonreconstruction} to produce meshes, a slower post-processing step that nevertheless results in significant noise and disconnected regions wherever predicted density is sparse---further demonstrating that representing amodal scenes as structured distance functions is essential for recovering physically plausible surface geometry. See Appendix~\ref{sec:supp_more_results} for more results.

\subsection{Ablation Study}

\begin{table}[t]
\centering
\small

\begin{minipage}[t]{0.38\linewidth}
    \centering
    \captionof{table}{\textbf{Metric geometry evaluation.}}
    \setlength{\tabcolsep}{3pt}
    \small
    \renewcommand{\arraystretch}{1.15}
    \begin{tabular}{lccc}
        \toprule
        \textbf{Method} & CD$\downarrow$ & FS$_{0.02}\uparrow$ & FPD$\downarrow$ \\
        \midrule
        LaRI            & 11.43   & 11.47 & --  \\
        NOVA3R & 6.56 & 22.23 & 16.50 \\
        \cdashline{1-4}
        \textbf{\Approach (ours)}   & \textbf{3.88} & \textbf{44.90} & \textbf{4.86} \\
        \bottomrule
    \end{tabular}
    \label{tab:metric_evaluation}
\end{minipage}%
\hfill
\begin{minipage}[t]{0.61\linewidth}
    \centering
    \caption{\textbf{VAE decoder design ablation.}}
    \setlength{\tabcolsep}{2pt}
    \small
    \renewcommand{\arraystretch}{1.15}
    \begin{tabular}{lcccc}
    \toprule
    \textbf{Decoder} &  L1 Dist$\downarrow$ & IoU$\uparrow$ & Time (s)$\downarrow$ & Mem. (GB)$\downarrow$ \\
    \midrule
    Dense-only & 0.49 & 93.2 & 0.63 & 9.2 \\
    Sparse-only & 0.48 & 93.0 & 0.09 & 2.5 \\
    \cdashline{1-5}
    \textbf{Hybrid (ours)} & \textbf{0.43} & \textbf{93.5} & \textbf{0.07} & \textbf{1.7} \\
    \quad w/ GT occupancy & 0.44 & 93.5 & 0.07 & 1.7 \\
    \quad w/ non-learned mask & 1.79 & 74.0 & 0.08 & 1.8 \\
    \bottomrule
    \end{tabular}

    \label{tab:vae_ablations}
\end{minipage}

    \vspace{-3mm}
\end{table}

\begin{figure}[t]
    \centering
    \begin{minipage}[c]{0.39\linewidth}
        \centering
        \captionof{table}{\textbf{Conditioning ablation.}}
        \setlength{\tabcolsep}{2pt}
        \small
        \renewcommand{\arraystretch}{1.15}
        \begin{tabular}{lcccc}
        \toprule
        \textbf{Config.} & $F_{\text{GFM}}$ & $z_{\text{vis}}$ & CD$\downarrow$ & FS$_{0.02}\uparrow$ \\
        \midrule
        
        \multirow{2}{*}{Single} & \xmark           &         \textit{Add}              & 6.80          & 31.72            \\
          & \textit{MoGe2}          & \xmark                     & 3.81           & 46.80            \\
        \cdashline{1-5}
        \multirow{2}{*}{Dual}         & \textit{MoGe2}          & \textit{Concat}           & 3.66           & 48.94            \\
         & \textit{MoGe2}         & \textit{Add}            & \textbf{3.03}  & \textbf{54.83}   \\
        \bottomrule
        \end{tabular}
        \label{tab:ablation_combined}
        \vspace{1.4em} %
        
        \centering
        \captionof{table}{\textbf{Foundation model ablation} (image-only conditioning).}
        \label{tab:image_conditioning_ablation}
        \setlength{\tabcolsep}{4pt}
        \small
        \renewcommand{\arraystretch}{1.15}
        \begin{tabular}{lcc}
            \toprule
            \textbf{Conditioning} & CD$\downarrow$ & FS$_{0.02}\uparrow$ \\
            \midrule
            DINOv2          & 5.46   & 35.98    \\
            VGGT            & 4.32   & 42.30    \\
            \cdashline{1-3}
            MoGe2           & \textbf{3.81} & \textbf{46.80} \\
            \bottomrule
        \end{tabular}
    \end{minipage}%
    \hfill
    \begin{minipage}[c]{0.6\linewidth}
        \centering
        \includegraphics[width=1\linewidth]{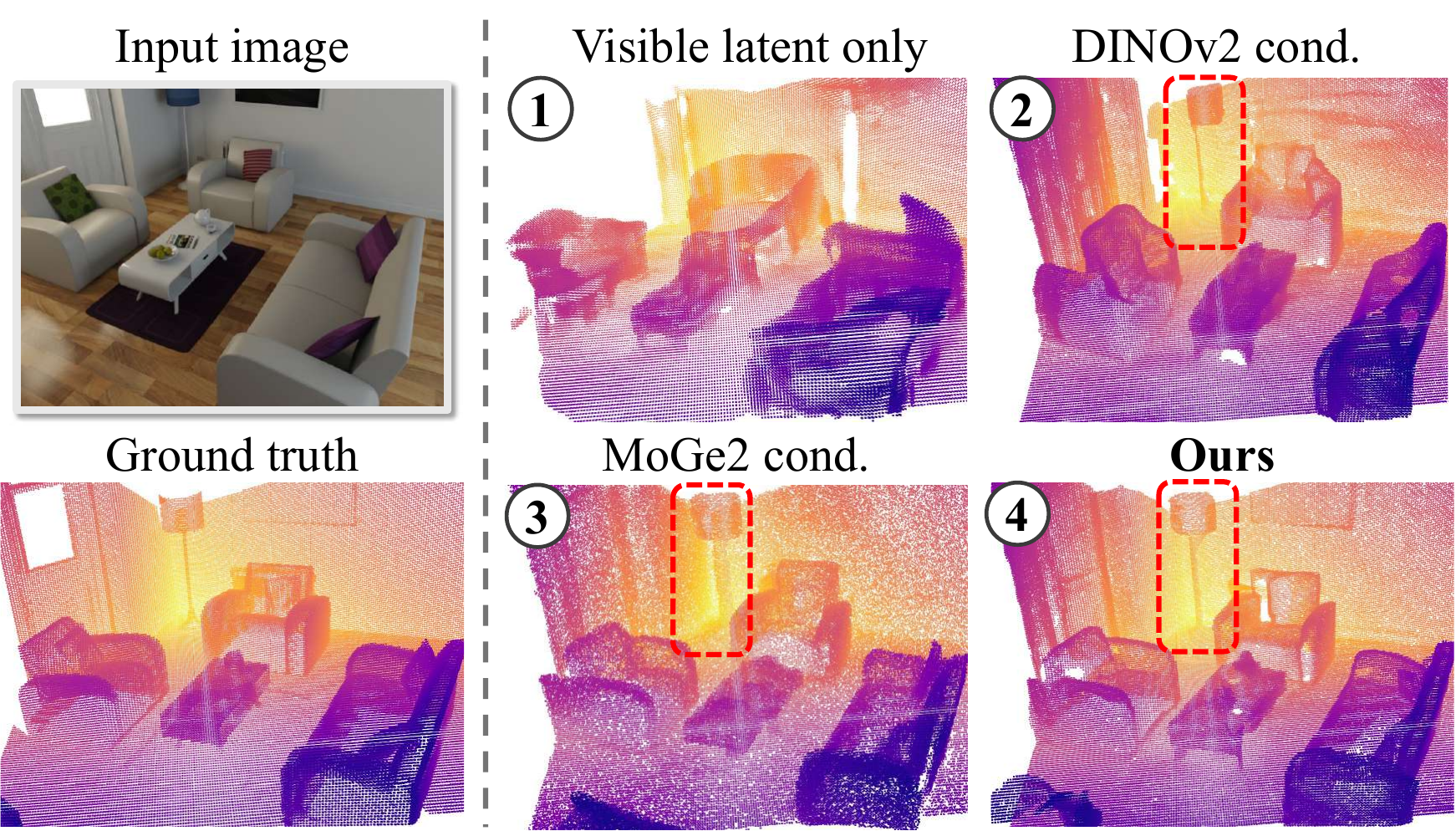}
        \caption{\textbf{Qualitative conditioning ablation.} \ding{172} Visible-only geometry fails to complete the scene; \ding{173}, \ding{174} image-only tokens yield distorted results; \ding{175} our dual-conditioning synthesizes sharp, high-fidelity amodal geometry.}
        \label{fig:abl_cond}
    \end{minipage}
    
\vspace{-5mm}
\end{figure}

We evaluate our core design choices through ablation studies on the SCRREAM dataset to justify our architectural components and conditioning strategies for effective amodal scene reconstruction.

\myheading{VAE decoder design} is analyzed in Table~\ref{tab:vae_ablations}. We evaluate reconstruction fidelity using the TUDF L1 distance within active cells and the occupancy Intersection over Union (IoU) from the thresholded TUDF. While a dense-only decoder is highly resource-intensive, a purely sparse variant reduces overhead. Both sparse variants share the same supervised occupancy head and differ only in the resolution at which it selects active voxels: filtering at the $16^3$ bottleneck expands each active cell into a coarse block at $256^3$, whereas our hybrid design upsamples densely to $64^3$ before filtering, isolating surface voxels more tightly. It thus achieves the best reconstruction quality alongside the lowest latency and memory footprint, validating its efficiency for compressing sparse 3D structures into a compact latent space. The supervised occupancy head is both near-optimal and essential: replacing its prediction with ground-truth occupancy leaves reconstruction unchanged (the head reaches $99.8\%$ occupancy IoU), whereas a non-learned mask collapses it (L1 $1.79$, IoU $74.0$).

\myheading{Ablation on visual and geometric priors} is shown in Table~\ref{tab:ablation_combined} and Fig.~\ref{fig:abl_cond}. Relying exclusively on the visible latent (\ding{172}) fails to recover unobserved structures. While image-only conditioning ($F_{\text{GFM}}$) enables amodal reasoning, using DINOv2 (\ding{173}) leads to distorted geometry, and replacing it with MoGe2 (\ding{174})---though more realistic---remains insufficient for recovering sharp, fine-grained details. Our full dual-conditioning strategy (\ding{175}) achieves the highest fidelity, where additive fusion via zero-initialized MLPs (\textit{Add}) outperforms concatenation (\textit{Concat}), demonstrating that the synergy of global visual context and explicit geometric grounding is essential for accurate amodal reconstruction.

\myheading{Ablation on different geometry foundation models} is analyzed in Table~\ref{tab:image_conditioning_ablation}. Despite its semantic richness, DINOv2 performs worst due to a lack of 3D spatial anchors. Geometry models like VGGT significantly improve results, while MoGe2 achieves the best performance. This suggests that structural consistency is a stronger driver for amodal reconstruction than general semantic reasoning.

\section{Limitations and Conclusion}
\label{sec:conclusion}

\myheading{Limitations.} Small holes can occasionally persist in heavily occluded regions, and since the scene is generated jointly, visible surfaces are slightly less sharp than those of MoGe2; reinstating the observed geometry (a training-free variant is evaluated in Appendix~\ref{sec:supp_reinstate}) or adding explicit 3D supervision are natural remedies. Our training and evaluation focus on indoor scenes; outdoor or very large scenes would benefit from more diverse training data and spatial chunking of the grid. We also fix the grid at $256^3$; since the decoder is largely sparse, its cost scales with the number of occupied voxels, so $512^3$ or $1024^3$ grids are feasible but require retraining the VAE and DiT, which we leave to future work. Finally, iterative sampling (1.4\,s on an RTX 4090) remains slower than single-pass regression.

\myheading{Conclusion.} We introduced \Approach, a generative framework for amodal 3D scene reconstruction. A hybrid 3D VAE compresses sparse TUDF grids into a compact latent space, while a latent Diffusion Transformer denoises this representation to recover the complete scene geometry; frozen geometric priors from MoGe2 provide the necessary spatial context. This shifts reconstruction from deterministic pixel-aligned regression to a structured generative manifold, enabling physically plausible completions in occluded regions. \Approach produces sharper geometry and outperforms existing amodal point-cloud and ray-based methods. 

\begin{ack}
Evangelos Kalogerakis has received funding from the European Research Council (ERC) under the Horizon research and innovation programme (Grant agreement No. 101124742).
\end{ack}

\clearpage
\bibliographystyle{plain}
\bibliography{neurips_2026}

\clearpage
\appendix

\section{Technical Appendices and Supplementary Material}
\label{sec:appendix}

\begin{figure}[t]
    \centering
    \includegraphics[width=0.98\linewidth]{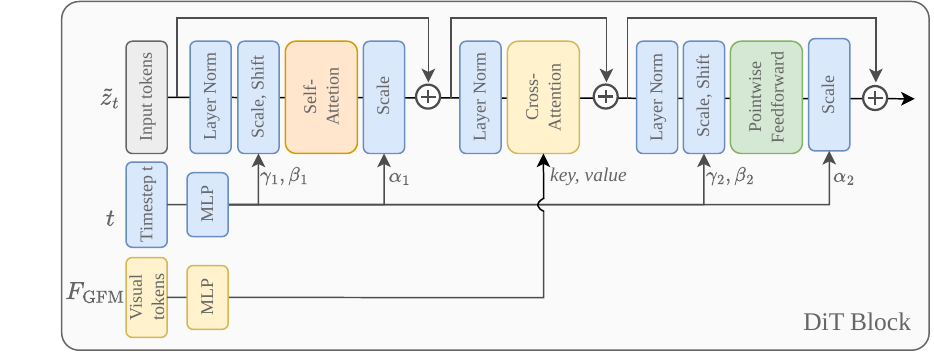}
    \caption{Illustration of the DiT block.}
    \label{fig:dit_block}
\end{figure}

\subsection{Volumetric Representation Design Choice}

We represent amodal 3D geometry as a Truncated Unsigned Distance Function (TUDF), where each voxel stores its distance to the nearest surface, clipped to a maximum of $\tau$ voxels. This choice is motivated by the limitations of the two alternatives below.

\emph{Binary occupancy} marks each voxel as occupied or empty, providing only a single bit of information per voxel, while TUDF encodes the continuous distance to the nearest surface. This continuous formulation provides smooth gradients that guide the decoder toward surface boundaries, whereas binary losses offer no directional signal within the truncation band. Additionally, TUDF enables precise sub-voxel surface localization through isosurface extraction, while binary occupancy is restricted to discrete voxel centers.

\emph{Truncated Signed Distance Functions (TSDF)} require a globally consistent inside/outside orientation, which is well-defined only for closed, watertight surfaces. Indoor scenes are frequently non-watertight (single-sided walls, floors, and furniture surfaces do not enclose a volume), so no such orientation exists. Consequently, TSDF provides no practical advantage over TUDF for open-geometry scenes. The unsigned formulation avoids this ambiguity entirely, as distance-to-nearest-surface remains well-defined regardless of scene topology.

\subsection{Architecture Details}
\label{sec:supp_arch}
Fig.~\ref{fig:dit_block} describes the internal components of our DiT block, which processes latent tokens $\tilde{z}_t$ through sequential self-attention, cross-attention, and pointwise feed-forward stages. To inject temporal information, the timestep $t$ is mapped via an MLP to generate adaptive layer normalization (AdaLN) parameters $\{\gamma_i, \beta_i, \alpha_i\}$ that modulate and scale  the features. Visual priors are integrated by passing foundation model tokens $F_{\text{GFM}}$ through an MLP to serve as key and value pairs for cross-attention. 

\subsection{Implementation Details}
\label{sec:supp_imp_details}
For the construction of the TUDF grids (Sec.~\ref{sec:problem_formulation}), the truncation distance $\tau$ is set to 3 voxels. In the VAE architecture (Sec.~\ref{sec:vae}), the bottleneck latent space ($16\times$ downsampling) utilizes a channel dimension of $C = 16$. To derive the low-resolution ground-truth binary occupancy mask from the high-resolution TUDF grid, we apply 3D max-pooling to accurately preserve the structural bounds of the geometry. Finally, the weighting coefficients for the composite VAE objective (Eq.~\ref{eqn:vae_loss}) are empirically balanced and set to $\lambda_{\text{bce}} = 0.2$, $\lambda_{\text{dice}} = 0.2$, and $\lambda_{\text{kl}} = 10^{-6}$.

\myheading{NRGB-D benchmark.} As described in Sec.~\ref{sec:exp}, we keep the NRGB-D frames whose fused geometry, cropped to the frame's view frustum, contains at least $0.2$ occluded voxels per visible voxel. Table~\ref{tab:supp_nrgbd} lists the per-scene breakdown of the resulting 137 samples.

\begin{table}[h]
\centering
\small
\caption{\textbf{NRGB-D evaluation set}: samples per scene after the amodal-enrichment filter.}
\label{tab:supp_nrgbd}
\setlength{\tabcolsep}{3pt}
\footnotesize
\begin{tabular}{lcccccccccc}
\toprule
Scene & breakfast & complete & green & grey-white & kitchen & morning & staircase & thin & white- & \textbf{Total} \\
 & room & kitchen & room & room & & apartment & & geometry & room & \\
\midrule
Samples & 17 & 15 & 12 & 14 & 15 & 13 & 20 & 11 & 20 & \textbf{137} \\
\bottomrule
\end{tabular}
\end{table}

\subsection{Training-Free Visible Reinstatement}
\label{sec:supp_reinstate}

Our visible surfaces are slightly less accurate than those of MoGe2 (Tables~\ref{tab:scrream_results} and~\ref{tab:nrgbd_results}) because \Approach generates the visible and hidden geometry jointly and conditions on the observed surface only softly, so it re-synthesizes and slightly smooths that surface instead of copying it (Sec.~\ref{sec:exp}). Since the observed geometry is already accurate, it can simply be kept. After decoding, we overwrite the predicted TUDF in the visible near-surface band, i.e., the voxels within $b=1$ voxel of the MoGe2 visible surface, with the visible TUDF computed from the MoGe2 pointmap, the same visible field the model is conditioned on. Both fields live on the same frustum-aligned $256^3$ grid by construction, so the merge is a per-voxel replacement with no resampling. Because only this one-voxel band changes, an occluded surface lying more than one voxel behind the visible surface is never affected. The step is training-free and optional; it is not used for the results in the main paper.

Table~\ref{tab:supp_reinstate} reports the result on both benchmarks under the evaluation protocol of Tables~\ref{tab:scrream_results} and~\ref{tab:nrgbd_results}; the \Approach rows repeat those tables. Reinstatement lowers the visible Chamfer distance to $2.01$ on SCRREAM and to $3.56$ on NRGB-D. Measured on identical samples, it also slightly improves occluded geometry (occluded CD $-0.27$ on SCRREAM and $-0.15$ on NRGB-D), since the more accurate visible points tighten the global Sim(3) alignment used for evaluation, while complete-scene accuracy changes little (complete CD $-0.10$ on SCRREAM and $-0.02$ on NRGB-D). We observed no seams between the reinstated band and the generated geometry.

\begin{table}[h]
\centering
\small
\caption{\textbf{Training-free visible reinstatement} on SCRREAM and NRGB-D ($\times 10^2$).}
\label{tab:supp_reinstate}
\setlength{\tabcolsep}{2.5pt}
\renewcommand{\arraystretch}{1.15}
\begin{tabular}{lccccccccc}
\toprule
\multirow{2}{*}{\textbf{Method}} & \multicolumn{3}{c}{\textbf{Visible}} & \multicolumn{3}{c}{\textbf{Occluded}} & \multicolumn{3}{c}{\textbf{Complete}} \\
\cmidrule(lr){2-4} \cmidrule(lr){5-7} \cmidrule(lr){8-10}
& CD$\downarrow$ & APD$_{0.1}\uparrow$ & APD$_{0.02}\uparrow$ & CD$\downarrow$ & APD$_{0.1}\uparrow$ & APD$_{0.02}\uparrow$ & CD$\downarrow$ & FS$_{0.1}\uparrow$ & FS$_{0.02}\uparrow$ \\
\midrule
\multicolumn{10}{l}{\textit{SCRREAM}} \\
\Approach & 2.84 & 96.10 & 56.73 & 3.46 & 92.70 & \textbf{55.53} & 3.03 & 95.03 & 54.83 \\
\, + visible reinstatement & \textbf{2.01} & \textbf{99.70} & \textbf{64.99} & \textbf{3.30} & \textbf{93.62} & 54.40 & \textbf{2.96} & \textbf{95.43} & \textbf{55.72} \\
\midrule
& CD$\downarrow$ & APD$_{0.1}\uparrow$ & APD$_{0.05}\uparrow$ & CD$\downarrow$ & APD$_{0.1}\uparrow$ & APD$_{0.05}\uparrow$ & CD$\downarrow$ & FS$_{0.1}\uparrow$ & FS$_{0.05}\uparrow$ \\
\midrule
\multicolumn{10}{l}{\textit{NRGB-D}} \\
\Approach & 4.52 & 90.06 & 69.22 & 6.97 & 81.41 & 62.48 & 6.44 & 84.35 & \textbf{63.92} \\
\, + visible reinstatement & \textbf{3.56} & \textbf{96.05} & \textbf{76.66} & \textbf{6.34} & \textbf{82.90} & \textbf{64.85} & \textbf{6.40} & \textbf{84.67} & 63.81 \\
\bottomrule
\end{tabular}
\end{table}

\subsection{Additional Quantitative Analysis}
\label{sec:supp_quant}

\myheading{Seed stability.} Tables~\ref{tab:scrream_results} and~\ref{tab:nrgbd_results} report a single sample per image. Table~\ref{tab:supp_seeds} repeats the SCRREAM evaluation with five sampling seeds, varying only the initial noise (all 460 samples, same protocol). The metrics change little across seeds (coefficient of variation $0.2$--$2.6\%$), and the single-seed values of Table~\ref{tab:scrream_results} lie close to the five-seed means, so the reported results are representative rather than a favorable draw.

\begin{table}[h]
\centering
\small
\caption{\textbf{Seed stability} on SCRREAM ($\times 10^2$): five sampling seeds (mean $\pm$ std) vs.\ the single seed of Table~\ref{tab:scrream_results}.}
\label{tab:supp_seeds}
\setlength{\tabcolsep}{4pt}
\renewcommand{\arraystretch}{1.15}
\begin{tabular}{lcccccc}
\toprule
 & Visible & \multicolumn{2}{c}{Occluded} & \multicolumn{3}{c}{Complete} \\
\cmidrule(lr){2-2} \cmidrule(lr){3-4} \cmidrule(lr){5-7}
 & CD$\downarrow$ & CD$\downarrow$ & APD$_{0.1}\uparrow$ & CD$\downarrow$ & FS$_{0.1}\uparrow$ & FS$_{0.02}\uparrow$ \\
\midrule
Single seed (Table~\ref{tab:scrream_results}) & 2.84 & 3.46 & 92.70 & 3.03 & 95.03 & 54.83 \\
Five seeds & 2.88\,$\pm$\,0.06 & 3.48\,$\pm$\,0.07 & 92.77\,$\pm$\,0.26 & 3.11\,$\pm$\,0.08 & 94.90\,$\pm$\,0.23 & 54.20\,$\pm$\,0.52 \\
\bottomrule
\end{tabular}
\end{table}

\myheading{Scene extent.} Our output grid is a fixed $256^3$ box fitted to the camera frustum, so its effective voxel size grows with the scene. To test whether this costs accuracy within the evaluated range, we split the 460 SCRREAM samples into tertiles by the grid's side length and report complete-scene accuracy per bin, under the protocol of Table~\ref{tab:scrream_results} (Table~\ref{tab:supp_extent}). Although the median voxel size nearly doubles from the small to the large bin ($8.7$ to $16.8$\,mm), accuracy does not drop for larger scenes: the large bin scores comparably to the medium one, while the small bin of close-range views is the hardest. Within this range, the $256^3$ grid does not limit accuracy; much larger scenes would still call for spatial chunking (Sec.~\ref{sec:conclusion}).

\begin{table}[h]
\centering
\small
\caption{\textbf{Accuracy vs.\ scene extent} on SCRREAM: complete-scene metrics per tertile of grid side length (CD and FS $\times 10^2$; median voxel size per bin).}
\label{tab:supp_extent}
\setlength{\tabcolsep}{5pt}
\renewcommand{\arraystretch}{1.15}
\begin{tabular}{lcccccc}
\toprule
Extent & Samples & Grid side (m) & Voxel size (mm) & CD$\downarrow$ & FS$_{0.1}\uparrow$ & FS$_{0.02}\uparrow$ \\
\midrule
Small  & 153 & 1.6--2.5 & 8.7  & 4.19 & 91.70 & 42.17 \\
Medium & 153 & 2.5--3.4 & 10.9 & 2.84 & 95.77 & 59.00 \\
Large  & 154 & 3.5--5.9 & 16.8 & 2.86 & 96.21 & 56.34 \\
\bottomrule
\end{tabular}
\end{table}

\subsection{Additional Qualitative Results}
\label{sec:supp_more_results}

\emph{We refer the reader to our project page,} \href{https://ngoductuanlhp.github.io/VolFill/}{\texttt{ngoductuanlhp.github.io/VolFill}}\emph{, for interactive 3D visualizations of the reconstructed scenes.}

\begin{figure}[t]
    \centering
    \includegraphics[width=0.98\linewidth]{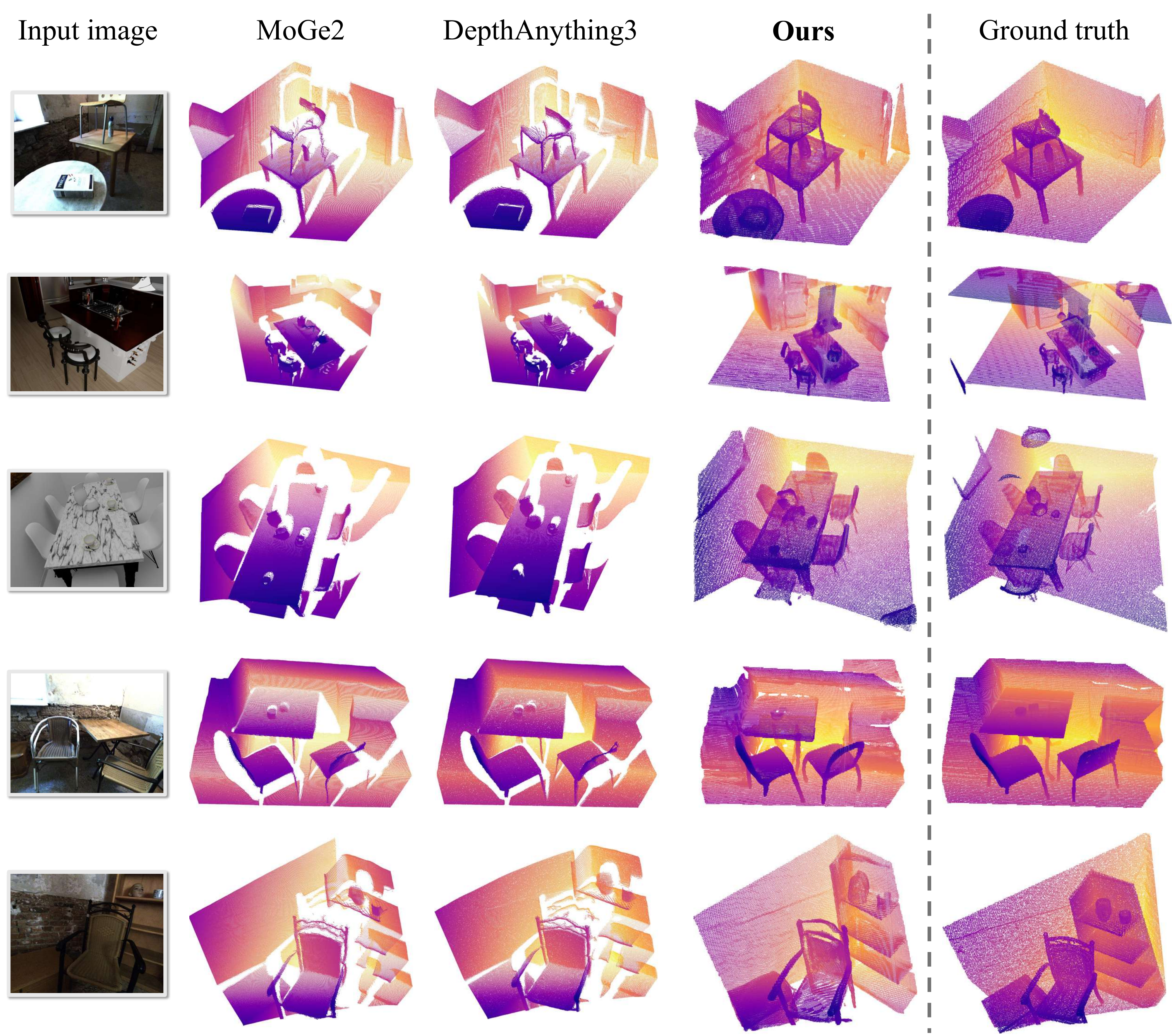}

    \caption{\textbf{Qualitative comparison with pixel-aligned approaches.} Unlike MoGe2~\cite{moge2} and DepthAnything3~\cite{depthanything3}, which are restricted to visible surfaces and leave significant holes, \Approach reconstructs complete, physically plausible amodal geometry.}    \label{fig:supp_quali_pixel_align}
\end{figure}

\myheading{Comparison with pixel-aligned geometry approaches} is presented in Fig.~\ref{fig:supp_quali_pixel_align}. While these baselines~\cite{moge2, depthanything3} achieve high accuracy on visible surfaces, they are inherently restricted to the camera's line of sight, leaving significant holes and ``shadows'' in occluded regions. In contrast, \Approach synthesizes physically plausible hidden structures, producing a continuous and structurally coherent scene representation resolving the visibility constraints of prior work.

\begin{figure}[t]
    \centering
    \includegraphics[width=0.98\linewidth]{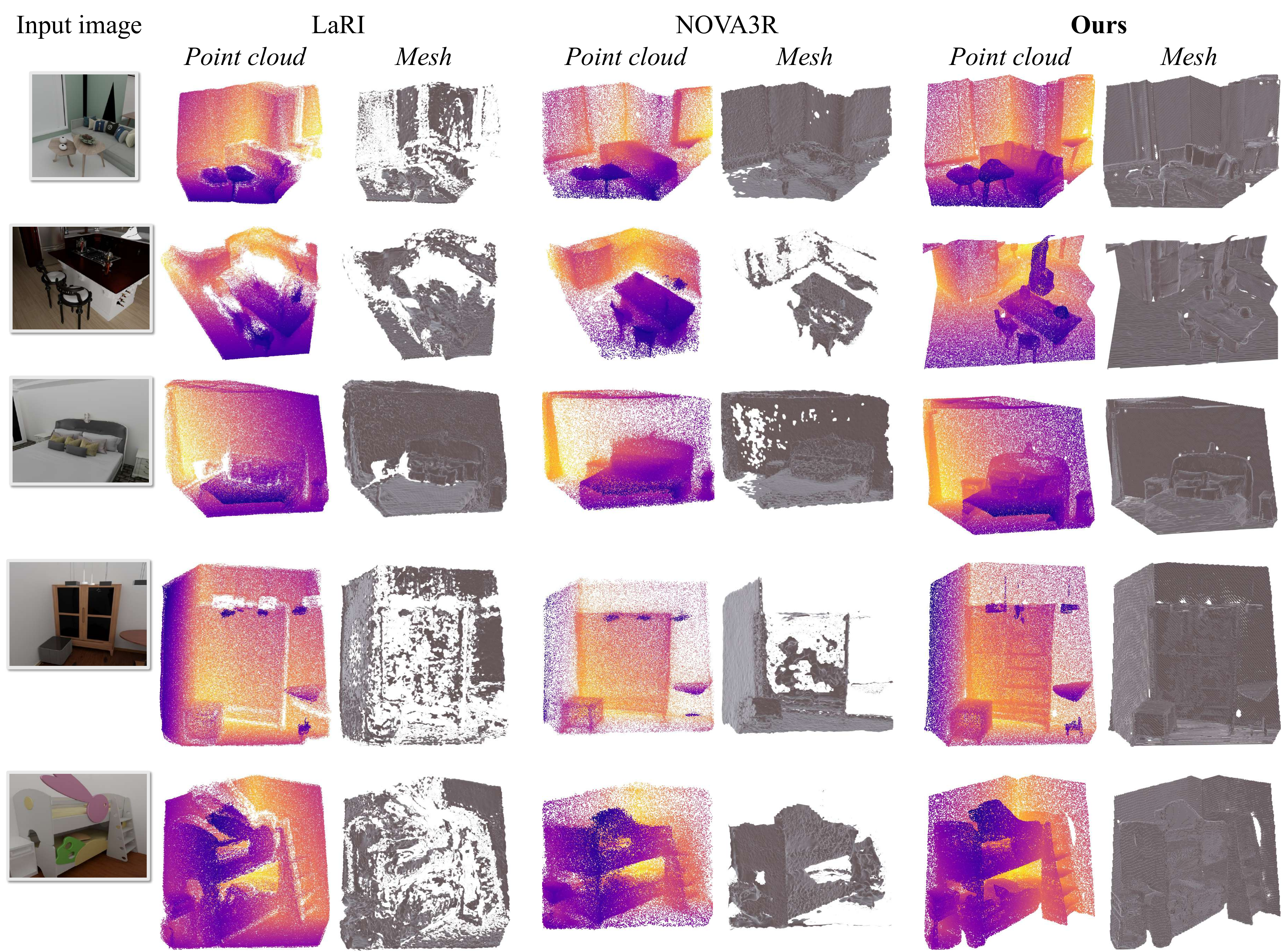}
    \caption{\textbf{Point cloud and mesh comparison.} Our method produces cleaner point clouds and significantly more coherent meshes than LaRI and NOVA3R.}    
    \label{fig:supp_quali_mesh}
\end{figure}

\myheading{Comparison of mesh reconstruction.} As shown in Fig.~\ref{fig:supp_quali_mesh}, LaRI and NOVA3R can produce reasonable point clouds in some cases, yet their unstructured outputs consistently yield fragmented, artifact-heavy meshes. Our approach produces cleaner point clouds and topologically consistent meshes that better reflect the physical scene geometry.

\begin{figure}[t]
    \centering
    \includegraphics[width=0.98\linewidth]{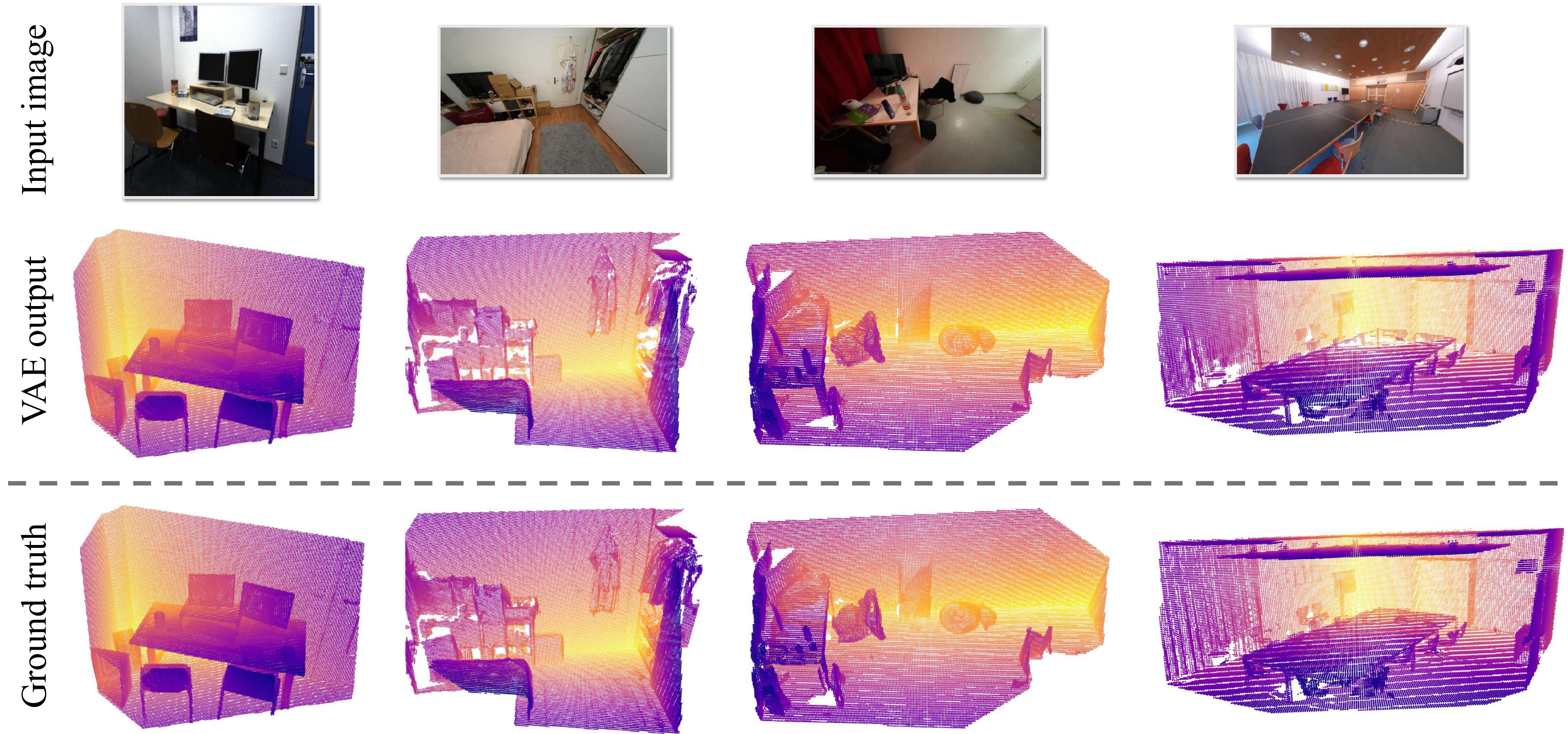}
    \caption{\textbf{Qualitative results} of the hybrid 3D VAE.}
    \label{fig:supp_quali_vae}
\end{figure}

\myheading{Qualitative results of the hybrid 3D VAE} are shown in Fig.~\ref{fig:supp_quali_vae}. Our VAE effectively compresses sparse high-resolution TUDF grids into a compact dense latent while preserving fine geometric details upon reconstruction. For larger scenes, a slight reduction in fidelity is observed given the fixed latent capacity, though overall structure and surface topology remain well-preserved---confirming that the latent space provides a faithful scene representation for the downstream diffusion process.

\begin{figure}[t]
    \centering
    \includegraphics[width=0.98\linewidth]{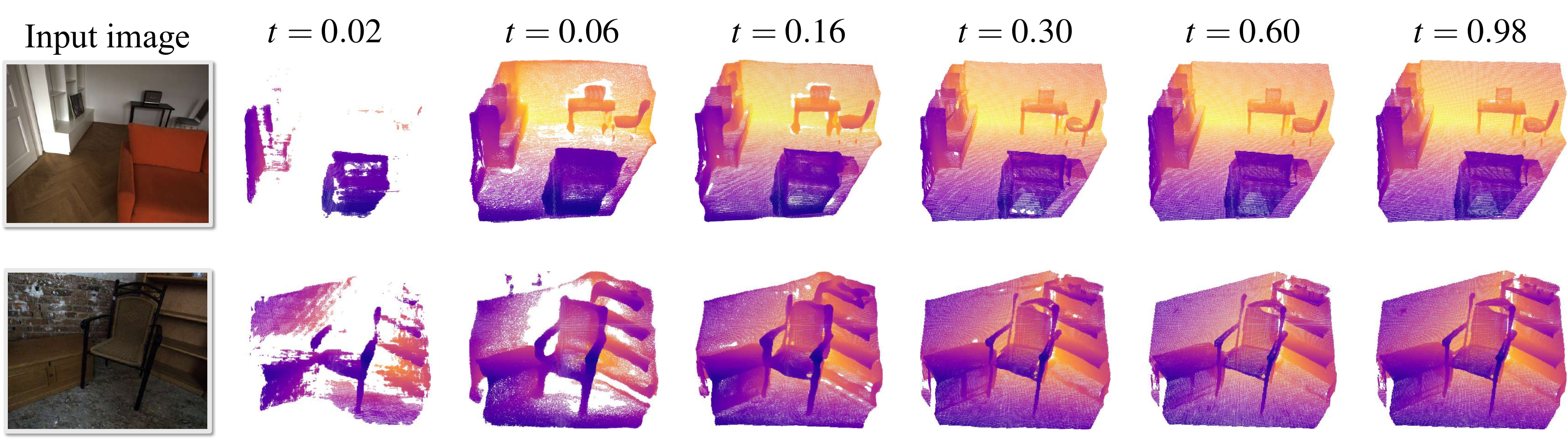}
    \caption{\textbf{Evolution of amodal geometry.} Clean-geometry estimates along the denoising trajectory at flow time $t \in (0,1)$ ($t=0$: noise, $t=1$: data). Our model rapidly converges on the coarse scene layout by $t=0.06$ and recovers sharp, detailed structures by $t=0.30$, followed by continuous refinement.}
    \label{fig:supp_denoise_traj}
 \end{figure}

\myheading{Generative inference trajectory.}
Fig.~\ref{fig:supp_denoise_traj} illustrates the iterative refinement of the amodal geometry throughout the denoising trajectory. Each panel decodes the model's clean-latent estimate at flow time $t$, which runs from $0$ (noise) to $1$ (data) over the $50$ Euler steps. Even at an early stage ($t=0.06$), the model already predicts the coarse overall structure of the scene. As the latent becomes progressively cleaner, the geometry sharpens significantly by $t=0.30$, followed by continuous, fine-grained refinement until the final step.

\begin{figure}[t]
    \centering
    \includegraphics[width=0.98\linewidth]{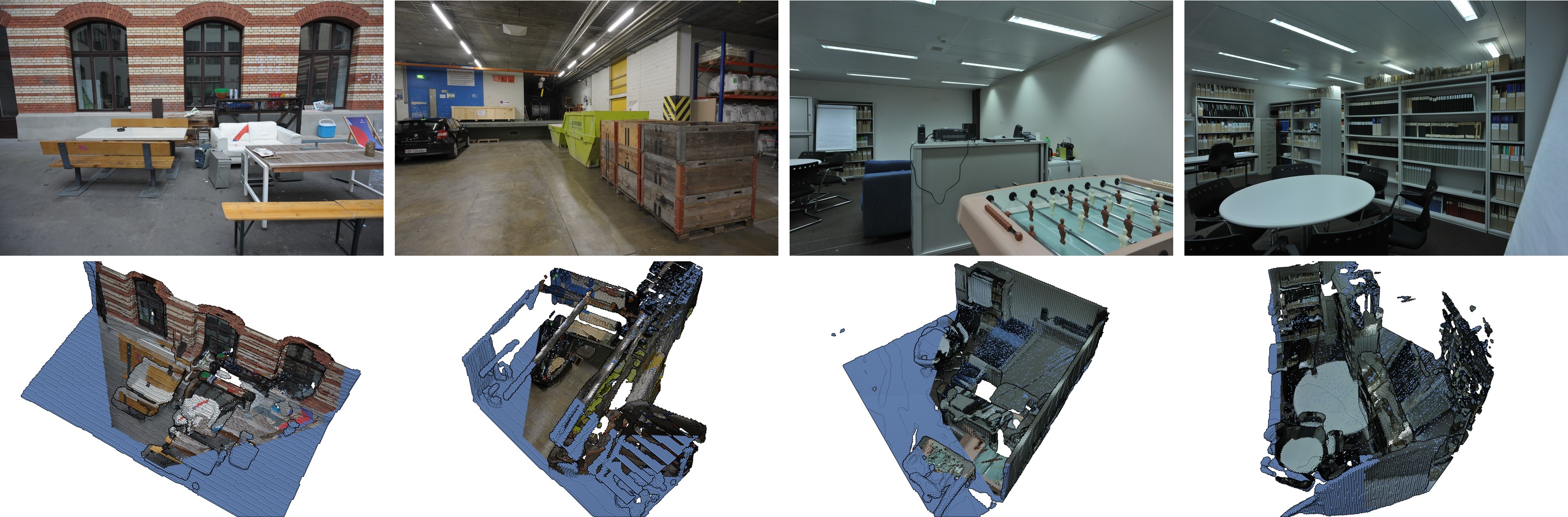}\\[1pt]
    {\small (a) Successes: ETH3D courtyard (semi-outdoor), warehouse, and two indoor office spaces.}\\[6pt]
    \includegraphics[width=0.98\linewidth]{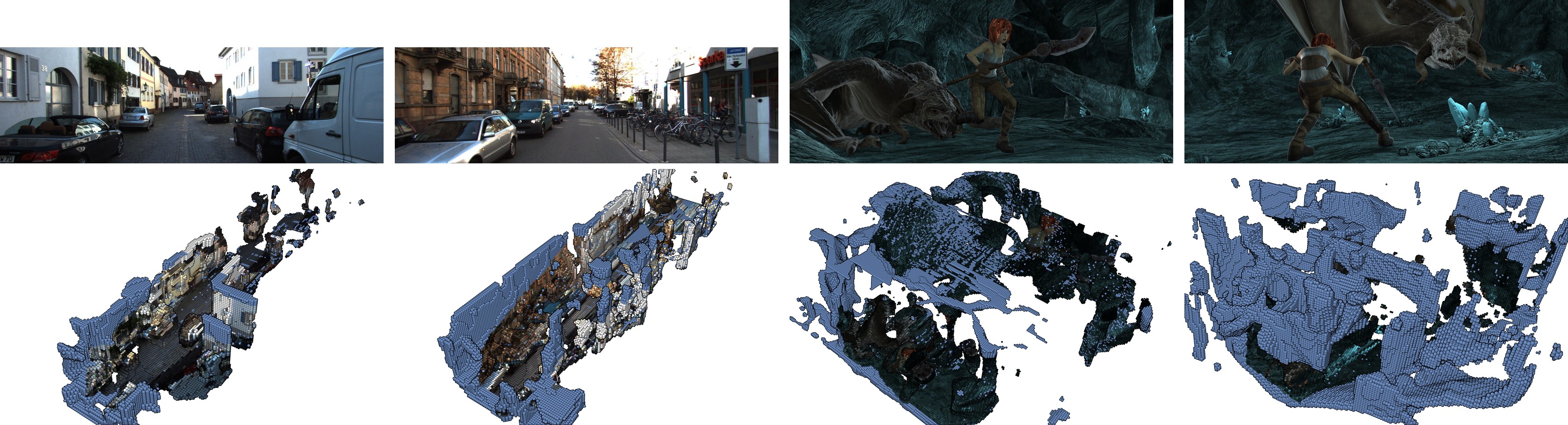}\\[1pt]
    {\small (b) Failures: KITTI street scenes and Sintel synthetic frames.}
    \caption{\textbf{Out-of-domain inputs.} \Approach, trained only on indoor data, applied without fine-tuning to images outside its training domain (input above each reconstruction). As in Fig.~\ref{fig:teaser}, geometry seen in the input view keeps the photo's color and hidden geometry recovered by \Approach is shown in \textcolor[HTML]{6E8FC4}{\textbf{blue}}.}
    \label{fig:supp_ood}
\end{figure}

\myheading{Out-of-domain inputs.} \Approach is trained only on indoor scenes (3D-FRONT and ScanNet++). Fig.~\ref{fig:supp_ood} shows its results, without any fine-tuning, on images outside this domain, including failure cases. On ETH3D~\cite{eth3d} (Fig.~\ref{fig:supp_ood}a), it still recovers coherent scene layouts with plausible hidden geometry for a semi-outdoor courtyard, a large warehouse (where the reconstruction becomes partly fragmented), and two indoor office spaces. It fails on long-range outdoor street scenes from KITTI~\cite{kitti} and on synthetic, low-light frames from Sintel~\cite{sintel} (Fig.~\ref{fig:supp_ood}b): the geometry fragments and the hidden regions become noisy. These scenes lie far outside the indoor training distribution in scale and appearance; extending \Approach to them would likely require more diverse training data and spatial chunking of the grid (Sec.~\ref{sec:conclusion}).

\section{Computing Resources}\label{sec:supp_compute}
Our research utilized multiple computing servers, consisting of a high-performance cluster for model training and a dedicated local workstation for data preparation, evaluation, and visualization.

\myheading{Hardware specifications.}
\begin{itemize}[leftmargin=1.5em]
\item \textbf{Cluster Training:} Primary training was conducted on a cluster node equipped with 2$\times$ NVIDIA RTX A6000 GPUs (48\,GB VRAM each). We also had access to a SLURM-managed server providing 2$\times$ NVIDIA L40S GPUs (48\,GB VRAM each).
\item \textbf{Local Workstation:} Data preprocessing, quantitative evaluation, and visualization were performed on a local workstation featuring an NVIDIA RTX 4090 GPU (24\,GB VRAM) and an Intel Core i9 processor with 64\,GB of RAM.
\end{itemize}

\myheading{Data preparation and training duration.}
The computational timeline for the final models reported in the main text is as follows:
\begin{itemize}[leftmargin=1.5em]
\item \textbf{Offline Preprocessing:} The preparation of the training data, including TUDF preparation and the pre-extraction of VAE latents to accelerate training, was performed on the local workstation. This process required approximately 3--4 days for the complete dataset.

\item \textbf{VAE Training:} The hybrid 3D VAE required approximately 3 days of training using 2 GPUs.
\item \textbf{DiT Training:} The latent Diffusion Transformer (DiT) required approximately 2.5 days using 2 GPUs.
\item \textbf{Ablation Studies:} The dense VAE ablation required approximately 6 days of compute, while DiT conditioning ablations mirrored the 2.5-day schedule.
\end{itemize}

\myheading{Total compute expenditure.}
Accounting for preliminary investigations and hyperparameter tuning, we estimate a total compute expenditure of approximately 1,680 GPU hours (2 GPUs over 35 days). This excludes the additional CPU-intensive preprocessing and evaluation time on the local workstation.

\section{Broader Societal Impacts}\label{sec:supp_impacts}

Our work advances single-image 3D scene reconstruction, which may benefit applications in robotics, assistive navigation, AR/VR, digital twins, architectural design, and embodied AI by enabling richer spatial understanding from limited visual input. In particular, complete-scene reconstruction could help autonomous systems reason about occluded geometry and improve safety in indoor navigation or human-robot interaction. However, the same capability may also raise concerns if deployed in privacy-sensitive settings, since inferred 3D layouts could reveal information about private homes, workplaces, or personal environments beyond what is directly visible in an image. The method could also be misused for surveillance, unauthorized mapping, or synthetic reconstruction of restricted spaces. In addition, models trained on datasets such as 3D-FRONT and ScanNet++ may inherit dataset biases toward particular indoor layouts, object categories, geographic regions, or socioeconomic settings, potentially limiting performance in underrepresented environments. Thus, we view this work as a research contribution rather than a deployment-ready system, and recommend that practical use include consent-aware data collection and privacy safeguards.

\clearpage
\newpage
\section*{NeurIPS Paper Checklist}

The checklist is designed to encourage best practices for responsible machine learning research, addressing issues of reproducibility, transparency, research ethics, and societal impact. Do not remove the checklist: {\bf The papers not including the checklist will be desk rejected.} The checklist should follow the references and follow the (optional) supplemental material.  The checklist does NOT count towards the page
limit. 

Please read the checklist guidelines carefully for information on how to answer these questions. For each question in the checklist:
\begin{itemize}
    \item You should answer \answerYes{}, \answerNo{}, or \answerNA{}.
    \item \answerNA{} means either that the question is Not Applicable for that particular paper or the relevant information is Not Available.
    \item Please provide a short (1--2 sentence) justification right after your answer (even for \answerNA). 
\end{itemize}

{\bf The checklist answers are an integral part of your paper submission.} They are visible to the reviewers, area chairs, senior area chairs, and ethics reviewers. You will also be asked to include it (after eventual revisions) with the final version of your paper, and its final version will be published with the paper.

The reviewers of your paper will be asked to use the checklist as one of the factors in their evaluation. While \answerYes{} is generally preferable to \answerNo{}, it is perfectly acceptable to answer \answerNo{} provided a proper justification is given (e.g., error bars are not reported because it would be too computationally expensive'' or ``we were unable to find the license for the dataset we used''). In general, answering \answerNo{} or \answerNA{} is not grounds for rejection. While the questions are phrased in a binary way, we acknowledge that the true answer is often more nuanced, so please just use your best judgment and write a justification to elaborate. All supporting evidence can appear either in the main paper or the supplemental material, provided in appendix. If you answer \answerYes{} to a question, in the justification please point to the section(s) where related material for the question can be found.

\begin{enumerate}

\item {\bf Claims}
    \item[] Question: Do the main claims made in the abstract and introduction accurately reflect the paper's contributions and scope?
    \item[] Answer: \answerYes{} %
    \item[] Justification: We clearly state the problem, contributions, and scope of our proposed method in both the abstract and introduction (Sec.~\ref{sec:intro}).

\item {\bf Limitations}
    \item[] Question: Does the paper discuss the limitations of the work performed by the authors?
    \item[] Answer: \answerYes{} %
    \item[] Justification: We discuss the limitations in Sec.~\ref{sec:conclusion}.

\item {\bf Theory assumptions and proofs}
    \item[] Question: For each theoretical result, does the paper provide the full set of assumptions and a complete (and correct) proof?
    \item[] Answer: \answerNA{} %
    \item[] Justification: We do not introduce theoretical results in the paper.

    \item {\bf Experimental result reproducibility}
    \item[] Question: Does the paper fully disclose all the information needed to reproduce the main experimental results of the paper to the extent that it affects the main claims and/or conclusions of the paper (regardless of whether the code and data are provided or not)?
    \item[] Answer: \answerYes{} %
    \item[] Justification: The method builds on the open-source MoGe2 model and is trained on public datasets (3D-FRONT and ScanNet++). Training, implementation, and evaluation details are described in Sec.~\ref{sec:exp} and Appendix~\ref{sec:supp_imp_details}. Thus, we believe that all the results can be reproduced.

\item {\bf Open access to data and code}
    \item[] Question: Does the paper provide open access to the data and code, with sufficient instructions to faithfully reproduce the main experimental results, as described in supplemental material?
    \item[] Answer: \answerNo{} %
    \item[] Justification: Inference code and pretrained checkpoints are publicly released; training, evaluation, and data-processing code, together with the processed datasets and split files, will also be released.

\item {\bf Experimental setting/details}
    \item[] Question: Does the paper specify all the training and test details (e.g., data splits, hyperparameters, how they were chosen, type of optimizer) necessary to understand the results?
    \item[] Answer: \answerYes{} %
    \item[] Justification: Training details, data splits, hyperparameters, evaluation datasets, and metrics are given in Sec.~\ref{sec:exp} and Appendix~\ref{sec:supp_imp_details}.

\item {\bf Experiment statistical significance}
    \item[] Question: Does the paper report error bars suitably and correctly defined or other appropriate information about the statistical significance of the experiments?
    \item[] Answer: \answerYes{} %
    \item[] Justification: Appendix~\ref{sec:supp_quant} (Table~\ref{tab:supp_seeds}) reports the mean and standard deviation over five sampling seeds on SCRREAM, varying only the initial noise; the variation is low. The main tables report one sample per image, aggregated over all test samples (Sec.~\ref{sec:exp}).

\item {\bf Experiments compute resources}
    \item[] Question: For each experiment, does the paper provide sufficient information on the computer resources (type of compute workers, memory, time of execution) needed to reproduce the experiments?
    \item[] Answer: \answerYes{} %
    \item[] Justification: We describe the computational resources used to train and evaluate our approach in Sec.~\ref{sec:exp} and in detail in Appendix~\ref{sec:supp_compute}.
    
\item {\bf Code of ethics}
    \item[] Question: Does the research conducted in the paper conform, in every respect, with the NeurIPS Code of Ethics \url{https://neurips.cc/public/EthicsGuidelines}?
    \item[] Answer: \answerYes{} %
    \item[] Justification: Our research conforms with the NeurIPS Code of Ethics, in every respect.

\item {\bf Broader impacts}
    \item[] Question: Does the paper discuss both potential positive societal impacts and negative societal impacts of the work performed?
    \item[] Answer: \answerYes{} %
    \item[] Justification:  We discuss positive and negative societal impacts in Appendix~\ref{sec:supp_impacts}.
    
\item {\bf Safeguards}
    \item[] Question: Does the paper describe safeguards that have been put in place for responsible release of data or models that have a high risk for misuse (e.g., pre-trained language models, image generators, or scraped datasets)?
    \item[] Answer: \answerNA{} %
    \item[] Justification: The released model reconstructs indoor 3D geometry from single images and poses no high risk of misuse; privacy-related concerns are discussed in Appendix~\ref{sec:supp_impacts}.

\item {\bf Licenses for existing assets}
    \item[] Question: Are the creators or original owners of assets (e.g., code, data, models), used in the paper, properly credited and are the license and terms of use explicitly mentioned and properly respected?
    \item[] Answer: \answerYes{} %
    \item[] Justification: We cite the original works for all datasets (3D-FRONT, ScanNet++, SCRREAM, Neural RGB-D, ETH3D, KITTI, Sintel) and models (e.g., MoGe2) used in the paper.

\item {\bf New assets}
    \item[] Question: Are new assets introduced in the paper well documented and is the documentation provided alongside the assets?
    \item[] Answer: \answerYes{} %
    \item[] Justification: We release inference code and pretrained checkpoints, documented on our project page; training, evaluation, and data-processing code, together with the processed data and split files, will be released with documentation.

\item {\bf Crowdsourcing and research with human subjects}
    \item[] Question: For crowdsourcing experiments and research with human subjects, does the paper include the full text of instructions given to participants and screenshots, if applicable, as well as details about compensation (if any)? 
    \item[] Answer: \answerNA{} %
    \item[] Justification: The paper does not involve crowdsourcing or research with human subjects.

\item {\bf Institutional review board (IRB) approvals or equivalent for research with human subjects}
    \item[] Question: Does the paper describe potential risks incurred by study participants, whether such risks were disclosed to the subjects, and whether Institutional Review Board (IRB) approvals (or an equivalent approval/review based on the requirements of your country or institution) were obtained?
    \item[] Answer: \answerNA{} %
    \item[] Justification: The paper does not involve crowdsourcing or research with human subjects.

\item {\bf Declaration of LLM usage}
    \item[] Question: Does the paper describe the usage of LLMs if it is an important, original, or non-standard component of the core methods in this research? Note that if the LLM is used only for writing, editing, or formatting purposes and does \emph{not} impact the core methodology, scientific rigor, or originality of the research, declaration is not required.
    \item[] Answer: \answerNA{} %
    \item[] Justification: LLMs are not a component of the core method; they were used only to assist with experiments (preprocessing datasets, writing visualization tools, implementing standard parts) and editing (e.g., grammar, spelling, word choice).

\end{enumerate}

\end{document}